\documentclass[journal]{IEEEtran}

\usepackage{graphicx}
\usepackage{amsmath}
\usepackage{amssymb}
\usepackage{algorithm}
\usepackage{algorithmic}

\usepackage{multirow}
\usepackage{enumitem}
\usepackage{makecell}
\usepackage{xspace}
\usepackage{verbatim}
\usepackage{rotating}
\usepackage[hidelinks,pagebackref,breaklinks,colorlinks=false,linkcolor=black]{hyperref}
\usepackage[capitalize]{cleveref}
\usepackage{subcaption}
\usepackage{url}
\usepackage[dvipsnames]{xcolor}

\graphicspath{{figures/}}

\newcommand{\ourapproach}{FedBNN\xspace}
\newcommand{\distD}{\mathcal{D}}

\crefname{figure}{Fig.}{Fig.}
\crefname{table}{Tab.}{Tab.}
\crefname{algorithm}{Alg.}{Alg.}

\begin{document}
	
\title{Variational Bayes for Federated Continual Learning}

\author{
	Dezhong~Yao,
	Sanmu~Li,
	Yutong~Dai,
	Zhiqiang~Xu,
	Shengshan~Hu,
 	Peilin~Zhao,
	and Lichao~Sun
	\IEEEcompsocitemizethanks{
		\IEEEcompsocthanksitem Dezhong Yao, Sanmu Li, and Shengshan Hu are with the National Engineering Research Center for Big Data Technology and System, Services Computing Technology and System Lab, Cluster and Grid Computing Lab, School of Computer Science and Technology, Huazhong University of Science and Technology, Wuhan 430074, China (E-mail: \{dyao, ltw\_cgcl, hushengshan\}@hust.edu.cn). Yutong Dai and Lichao Sun are with Lehigh University, PA 18015, USA (E-mail: \{yud319, lis221\}@lehigh.edu). Zhiqiang Xu is with Mohamed bin Zayed University of Artificial Intelligence, UAE (E-mail: zhiqiang.xu@mbzuai.ac.ae). Peilin Zhao is with Tencent AI Lab, Shenzhen, China. (E-mail: masonzhao@tencent.com)
	}
	\thanks{\protect (Corresponding author: Dezhong Yao), Under review. }
}



\maketitle

\IEEEtitleabstractindextext{
\begin{abstract}
	Federated continual learning (FCL) has received increasing attention due to its potential in handling real-world streaming data, characterized by evolving data distributions and varying client classes over time. The constraints of storage limitations and privacy concerns confine local models to exclusively access the present data within each learning cycle. Consequently, this restriction induces performance degradation in model training on previous data, termed ``catastrophic forgetting''.
	However, existing FCL approaches need to identify or know changes in data distribution, which is difficult in the real world. 
	To release these limitations, this paper directs attention to a broader continuous framework.
	Within this framework, we introduce Federated Bayesian Neural Network (\ourapproach), a versatile and efficacious framework employing a variational Bayesian neural network across all clients.  
	Our method continually integrates knowledge from local and historical data distributions into a single model, adeptly learning from new data distributions while retaining performance on historical distributions. 
	We rigorously evaluate \ourapproach's performance against prevalent methods in federated learning and continual learning using various metrics. Experimental analyses across diverse datasets demonstrate that \ourapproach achieves state-of-the-art results in mitigating forgetting.
\end{abstract}


}

\IEEEdisplaynontitleabstractindextext

\IEEEpeerreviewmaketitle

\vspace{3.0em}

\IEEEraisesectionheading{\section{Introduction}}
	
\IEEEPARstart{F}{ederated} Learning (FL) facilitates collaborative training of a global model among mobile devices or small-scale organizations, effectively overcoming data silos and data privacy challenges~\cite{flsurvey2020Li, fang2022robust}. Serving as a communication-efficient and privacy-preserving training scheme, FL has been widely used in various applications, such as keyboard prediction~\cite{flkeyboard2018Hard, ramaswamy2019federated}, medical diagnosis~\cite{yang2021flop}, autonomous driving~\cite{samarakoon2019distributed}, and object detection~\cite{nguyen2019diot,li2024fuse}.

\begin{figure}[t]
    \centering
    \begin{subfigure}[t]{0.98\linewidth}
        \centering
        \includegraphics[width=\textwidth]{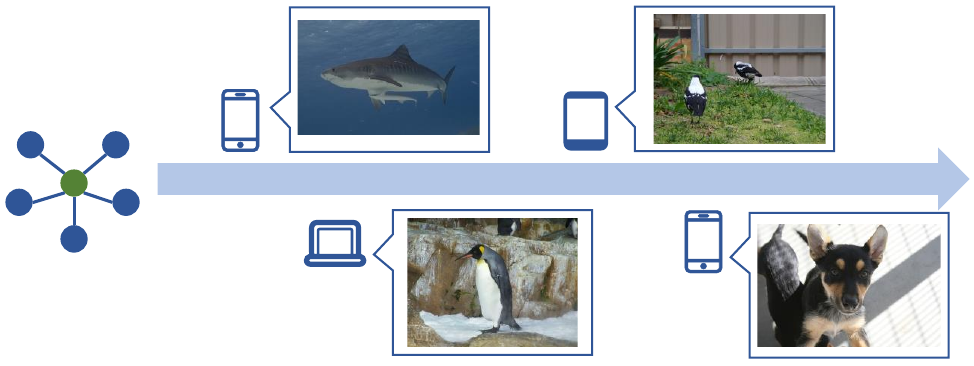}
        \caption{Federated class-incremental: users report new classes in a federated object detection system.}
        \label{fig:class_incr}
    \end{subfigure}
    \begin{subfigure}[t]{0.98\linewidth}
        \centering
        \includegraphics[width=\textwidth]{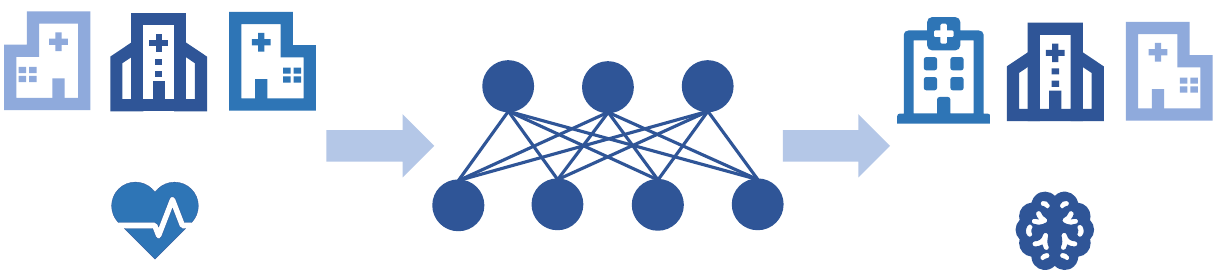}
        \caption{Federated task-incremental: a federated diagnosis system switch from the cardiology task to the neurology task.}
        \label{fig:domain_incr}
    \end{subfigure}
    \caption{In many real world scenarios, the data distribution of a federated learning system will evolve over time~\cite{gong2022ode}. Therefore, approaches that deal with dynamic data distribution are desirable for federated learning systems.}
    \vspace{-1.5em}
    \label{fig:dynamic_task}
\end{figure}

However, traditional FL algorithms operate under the assumption that data distribution and classes remain constant across all devices. Real-world scenarios, however, witness clients continually encountering new concepts and evolving data distribution over time~\cite{fcil2022Dong, tang2021layerwise}. As illustrated in \cref{fig:dynamic_task}, instances such as the emergence of new object detection classes reported by individual users (\cref{fig:class_incr}), or the application of a federated diagnostic system to a new department within a healthcare institute (\cref{fig:domain_incr}), exemplify the evolving nature of data. Local clients, constrained by limited storage and privacy considerations, typically retain recent data, granting local models access solely to current data during new learning cycles. This leads to two problems: 1) rapid adaptation of local models to recent classes results in significant degradation in performance on previously learned distribution, known as the \textit{catastrophic forgetting}~\cite{cf1989mccloskey,cf2019Benedikt}, and 2)  a set of local models in FL exhibit performance drop on disparate tasks, referred to as the \textit{negative knowledge transfer}~\cite{das2022federated}. These challenges render the use of conventional FL in a continual setting inefficient. This motivates us to extend the conventional FL to deal with the federated continual tasks that follow a non-stationary distribution, so we can dynamically update the model to effectively exchange knowledge while avoiding forgetting.

\IEEEpubidadjcol
	
Recently, some efforts have been made to solve federated continual learning challenges stemming from non-stationary distribution. For example, online learning methods~\cite{liu2020fedvision,gong2022ode} minimize the accumulative loss from the perspective of regret minimization. These algorithms require buffering previous data for training, which may be unrealistic as the size of datasets often prohibits frequent batch updating. Also, continual-learning-based approaches~\cite{fedweit2021yoon, fcil2022Dong} avoid revisiting previous data and aim to overcome catastrophic forgetting. These approaches require identification~\cite{fcil2022Dong} or awareness of data distribution changes~\cite{fedweit2021yoon}, which is also unrealistic when there is no clear task boundary.

In this paper, we propose a Bayesian neural network (BNN) based approach, named \textbf{\ourapproach}\footnote{The source code is available at \protect\url{https://github.com/LasSino/fedbnn_code}.}, which tackles the issues of negative knowledge transfer and catastrophic forgetting for general FCL tasks.
By integrating knowledge of previous data into the BNN prior and properly aggregating locally learned BNN distributions, \ourapproach demonstrates the capability to sustain its performance on past data distributions while continually learning from new distributions, despite the data heterogeneity among clients. Remarkably, the proposed approach is task-agnostic by nature and can be applied to the general FCL problem. 
It is empirically shown that the Bayesian formulation of our algorithm effectively alleviates negative transfer among different tasks to overcome the catastrophic forgetting issue.
	
\noindent\textbf{Summary of Contributions:} 
\begin{itemize}
    \item We formulate a comprehensive definition of FCL, including both class-incremental and task-incremental scenarios, which do not require clear task boundaries.
    \item We propose \ourapproach, the first variational-based FCL method, to the best of our knowledge. Unlike the previous works, \ourapproach can handle FCL without obvious task boundaries.
    \item We verify the effectiveness of our proposed method over existing baselines via extensive experiments on various FCL settings.
\end{itemize}

\section{Backgrounds}
	
\subsection{Related Work}

\textbf{Continual Learning}  tries to train a single model that can learn from multiple tasks sequentially \cite{clsurvey2019Parisi,clsurvey2022lange,yan2022learning,wang2021ordisco}. 
Previous works can be categorized in terms of the continual learning setting as follows: class-incremental learning, task-incremental learning, and domain-incremental learning. Many practical approaches have been proposed to deal with specific settings, like iCaRL \cite{icarl2017Rebuffi}, EWC \cite{ewc2017kirkpatrick}, and LwF \cite{LwF2016Li}. These settings assume there are well-defined boundaries between tasks, and many of them require the boundary known to the algorithm. However, recent advances introduce task-agnostic continual learning, which dispenses with the need for clear task boundary, and assumes task transitions happen gradually \cite{tagcl2019He,tagcl2021Zeno}. This is a more realistic and general setting of dynamic learning tasks. In this study, we adopt this setting, reflecting the evolving nature of learning tasks over time.
	
\textbf{Federated Continual Learning} has emerged to learn new tasks continuously while tackling forgetting on old tasks in federated scenarios~\cite{fclsurvey2022Criado,realworldfcl2022Dupuy}.
Some approaches have been proposed to learn a global continual model among clients. CFeD~\cite{CFeD2022Ma} overcomes catastrophic forgetting on clients by knowledge distillation. FCIL~\cite{fcil2022Dong, fcil24dong} focuses on federated class-incremental settings, addressing catastrophic forgetting by mitigating the imbalance in sample classes. FISS~\cite{fiss2023Dong} targets the specific application of federated incremental semantic segmentation via adaptive class-balanced pseudo labeling. Notably, these approaches require identifying or knowing changes in data distribution, which is unrealistic in many real-world scenarios.
Another series of work focuses on helping each client continuously learn its local model with indirect knowledge from other clients. FedWEIT~\cite{fedweit2021yoon} achieves this objective by decomposing the model into global parameters and task-specific parameters. FedKNOW~\cite{fedknow2023Luopan} further enhances communication efficiency over FedWEIT by storing more lightweight signature knowledge for each local task. In this paper, we still focus on learning a global continual model, since a global model is desirable in prediction on server and newly joined clients \cite{gfl2023Chen,gpfl2023Zhang}.

\begin{figure*}[t]
    \centering
    \begin{subfigure}[t]{0.31\linewidth}
        \centering
        \includegraphics[width=\textwidth]{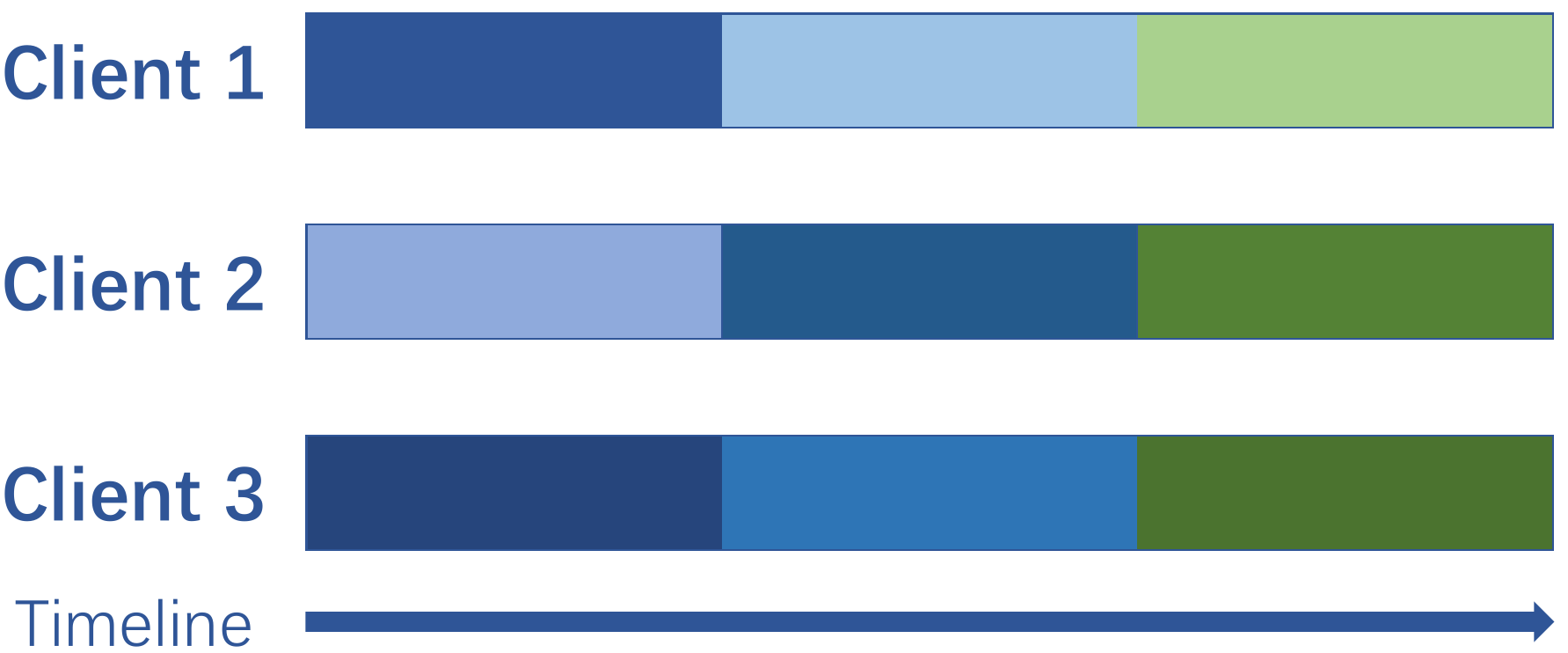}
        \caption{All clients change to a new distribution simultaneously. Note the heterogeneity among clients on the same task.}
        \label{fig:fcl_case_1}
    \end{subfigure}
    \hspace{0.02\linewidth}
    \begin{subfigure}[t]{0.31\linewidth}
        \centering
        \includegraphics[width=\textwidth]{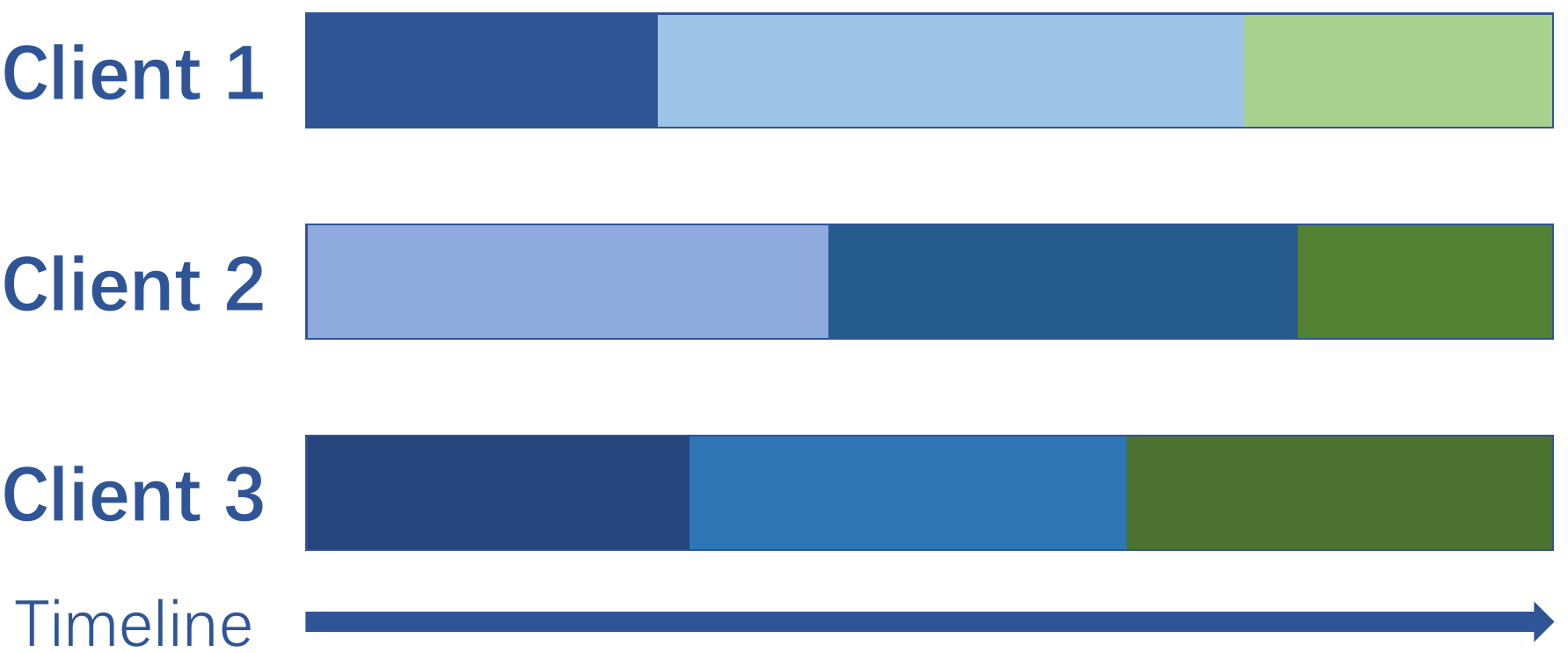}
        \caption{Some clients change to a new distribution ahead of others.}
        \label{fig:fcl_case_2}
    \end{subfigure}
    \hspace{0.02\linewidth}
    \begin{subfigure}[t]{0.31\linewidth}
        \centering
        \includegraphics[width=\textwidth]{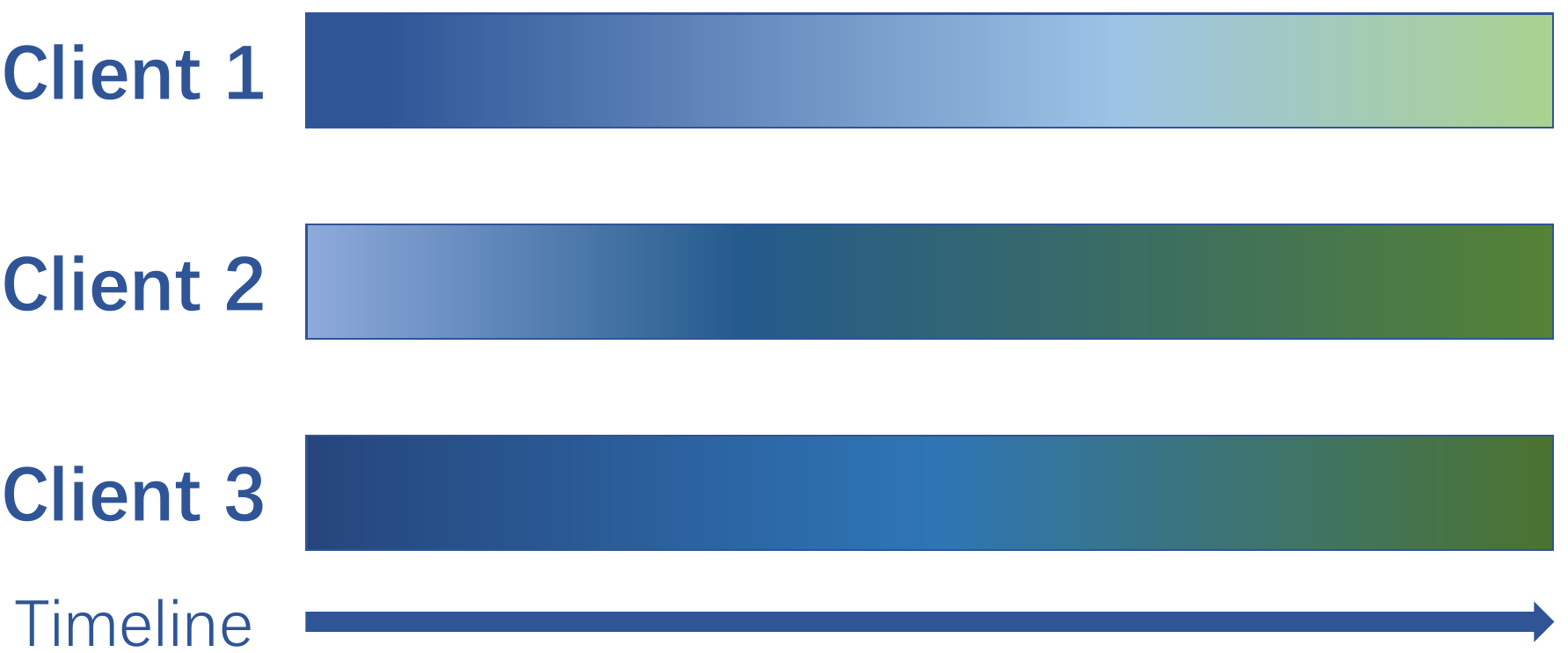}
        \caption{Distribution on each client drifts gradually at different speed.}
        \label{fig:fcl_case_3}
    \end{subfigure}
    \vspace{-0.5em}
    \caption{In real world FCL applications, data evolution on clients can exhibit different patterns. The figure demonstrates three typical cases of FCL data distribution. Difference of data distribution is demonstrated by different colors. }
    \vspace{-1.0em}
    \label{fig:fcl_cases}
\end{figure*}

\textbf{Bayesian Neural Network} provides a novel probabilistic perspective of neural networks~\cite{bnn1996Neal,bnn2015blundell}. It views network parameters as random variables and solves for the posterior distribution of the parameters. Many approaches have been proposed to obtain the posterior distribution, including Expectation Propagation (EP)~\cite{bnn_ep01Minka, bnn_ep20Zhao}, Variational Inference (VI)~\cite{bnn2015blundell, vi19zhang, bnn_vi23Hubin}, and Markov Chain Monte Carlo (MCMC)~\cite{bnn_mcmc00Vehtari, bnn_mcmc19Li, bnn_mcmc21Li}. Recently, BNN-based approaches have been used in FL~\cite{bnn_fl21AlShedivat, bnn_fl22Kassab, bayesfl24liu} and CL~\cite{vcl18nguyen, bgd2018Zeno, bnn_cl21Loo}, and show promising results in solving specific challenges in FL and CL scenarios. 
	
\subsection{Preliminary: BNN and VCL}
Bayesian Neural Network (BNN) is an extension of a standard neural network by incorporating Bayesian principles. In BNN, the parameters are treated as random variables. Therefore, instead of a single value, model parameters $\theta$ in BNN are modeled by some distribution $p(\theta)$. Given a dataset $\distD$, the fundamental objective in BNN learning is to find the conditional distribution of model parameters on given data $p(\theta | \distD)$, known as the posterior. Specifically, the Bayes rule is applied: 
\begin{eqnarray}
    p(\theta | \distD) = \frac{p(\theta) p( \distD | \theta) }{p( \distD)} \propto p(\theta) p( \distD | \theta)
    \label{eqn:bayesian_rule_of_param}
\end{eqnarray}
The first term is a prior distribution for the parameter, while the latter is the data-related likelihood. This distribution is intractable itself, but it is viable to approximate it by some parameterized distribution families $q(Z)$, known as variational inference.  
The notable work of Bayes by Backprop \cite{bnn2015blundell} enables efficient inference on large BNNs. It uses mean-field Gaussian distribution as the variational distribution, and makes the approximation via backpropagation and gradient descent, in a manner similar to standard neural networks.
	
Given the above property of BNN, Variational Continual Learning~(VCL) \cite{vcl18nguyen} was proposed to address the challenge of continual learning. After a sequence of tasks  $ \distD_{1 \ldots T} = \{ \distD_1,\distD_2,\ldots , \distD_T \}$, the desired parameter distribution $p(\theta | \distD_{1 \ldots T})$ can be factorized as follows.
	\begin{eqnarray}
		p(\theta | \distD_{1 \ldots T}) \propto p(\theta) p( \distD_{1 \ldots T} | \theta)
	\end{eqnarray}
	It is assumed that the dataset samples are conditional independent given parameter $\theta$, so the probability $p( \distD_{1 \ldots T} | \theta)$ is multiplicative, i.e. 
	\begin{eqnarray}
		p( \distD_{1 \ldots T} | \theta) = \prod_{t=1}^{T}{p( \distD_t | \theta)}
		\label{eqn:vcl_likelihood_decomp}
	\end{eqnarray}
	Therefore, the following recurrence relation holds:
	\begin{eqnarray}
		\label{eqn:vcl}
		\begin{aligned}
			p(\theta | \distD_{1 \ldots T}) 
			& \propto p(\theta) \prod_{t=1}^{T}{p( \distD_t | \theta)} \\
			& \propto p(\theta | \distD_{1 \ldots T-1}) p( \distD_T | \theta)
		\end{aligned}
	\end{eqnarray}
	VCL utilizes the relation to address the problem of continual learning. When a new task is encountered, a new posterior is obtained by setting the latest posterior as the prior and then performing variational inference on the new task. Consequently, the new posterior integrates knowledge about past tasks while learning from the new task, thus robust to distribution change in task shifts.
	
	The success of VCL offers valuable insight into BNN's capacity to integrate knowledge of multiple data distributions into a single model adaptively and continuously. Therefore, employing BNN can be a viable approach to address the FCL problem.
	
\section{Problem Formulation of FCL}
	It is complicated for multiple clients to collaboratively train a global model from dynamic data streams.
	To better understand the federated continual learning problem, we first analyze the three main characteristics of real-world FCL scenarios.
	Based on the analysis, we propose a generalized problem definition that closely fits real-world cases.
	
\subsection{Motivation: Real World FCL Scenarios}
	
	
	Firstly, FCL systems are expected to operate over extended periods of time, during which the client data evolves. In addition, since training data are generated locally by clients, FCL also needs to handle heterogeneous data among clients (known as the non-IID problem). The primary objective of FCL is to collaboratively train a global model that continuously learns from new data on clients while preserving its performance on previous data distribution. 
	
	
	Secondly, the task boundary of the data stream on each client is unknown, and the data evolution among clients can exhibit various patterns. \cref{fig:fcl_cases} illustrates three potential cases of FCL data distribution. In the first case \cref{fig:fcl_case_1}, the whole FCL system may change its learning data simultaneously and instantly. Alternatively, in the second case \cref{fig:fcl_case_2}, some clients experience data distribution change before others. Finally, the third case \cref{fig:fcl_case_3} demonstrates a gradual drift in client distributions over time, each progressing at a distinct rate. This asymmetric evolution of data distribution is commonly observed in large-scale FCL scenarios. It is important to note that there are typically \textit{no clear task boundaries} during this evolution process, and even if such boundaries exist, they often remain undisclosed to the learning system \cite{tagcl2019He}.
	
	
	Thirdly, while the distribution changes occur separately for each client, we make the assumption that there is a \textit{global trend} followed by the drift in client distributions.
	This assumption forms the prerequisite for FCL systems. If the distribution drift among clients diverges significantly, collaborative learning among the clients becomes ineffective, rendering FCL approaches unsuitable. Fortunately, this assumption is generally satisfied since trending is a prevalent social phenomenon.
	
    In summary, under real-world FCL scenarios, we need an approach that (1) deals with distributed and evolving data distribution, (2) can be applied to various FCL cases, and (3) is task-agnostic by nature.
	
\subsection{Problem Definition}
\noindent\textbf{General Federated Continual Learning}: Suppose $K$ clients collaborate to train a machine learning model, parameterized by $\theta$. Each client $k$ works on its local stream data, and at time $t$, the local data samples $(X, Y)$ follow some distribution $ (X, Y) \sim \distD_k^t$. At  time $t$, distributions on clients are different from each other: $\exists k_1 \neq k_2 : \distD_{k_1}^t \neq \distD_{k_2}^t$. Meanwhile, distribution on the same client will change over time: $\exists t_1 \neq t_2: \distD_k^{t_1} \neq \distD_k^{t_2}$.
	
Although client data is always stored locally, at time $t$, all clients' data forms a global distribution $ \bigcup_{k} {\distD_k^t}=\distD^t $. Due to the evolution of client data, the global distribution is also dynamic: $ \exists t_1 \neq t_2: \distD^{t_1} \neq \distD^{t_2}$.
	

	At the time $t$, only a batch of local data from the current distribution $D_k^t$ is accessible to client $k$. The goal of FCL is to learn a new global parameter $\theta^t$ based on the latest global parameter $\theta^{t-1}$ that minimizes loss on all the historical distributions. The objective function at time $t_c$ can be expressed as:
	\vspace{-0.5em}
	\begin{eqnarray} \label{eqn:fcl_obj}
		\theta^{t_c} = 
		\mathop{\arg\min}\limits_{\theta} 
		\sum_{t=0}^{t_c} \mathcal{L}(\theta;\distD^t) =
		\mathop{\arg\min}\limits_{\theta} \mathcal{L}(\theta;\distD^{1 \ldots t_c})
	\end{eqnarray}
	where $\mathcal{L}(\theta;\distD)$ is the loss function on data distribution $\distD$.
	
	This formulation characterizes the client heterogeneity and temporal dynamic of the FCL problem, aiming to learn a model that minimizes the loss of all past data. This formulation is general enough for various real-world scenarios, encompassing the range of cases discussed earlier.

 \begin{figure*}[t]
    \centering
    \includegraphics[width=\linewidth]{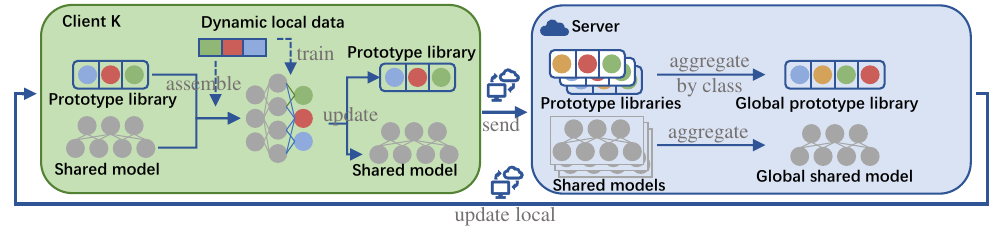}
    \vspace{-1.2em}
    \caption{The mechanism of prototype library in \ourapproach. Before local training, the classifier layer is assembled according to classes of current local data. The classifier layer is appended to the shared model for training. After training, the prototypes in the classifier layer are used to update the local prototype library. Local prototype libraries are sent and aggregated by class on the server, then sent back to clients.}
    \label{fig:prototype_lib}
    \vspace{-1.0em}
\end{figure*}

	\section{The Proposed \ourapproach Approach}
	\label{sec:fedbnn}
	To deal with dynamic and heterogeneous data streams in FCL, we propose \ourapproach. Based on Bayesian Neural Network, our approach is designed to work with the general FCL problem in the real world. 
	
	
	\subsection{Federated Bayesian Neural Network}
	Inspired by BNN's potential in learning continuously from evolving data distribution, we propose to address the problem of federated continual learning by utilizing a Bayesian neural network. However, the distributed nature of data across numerous clients, coupled with the continual evolution of data distributions among these clients, renders direct learning of a global BNN model unfeasible within the context of FCL. To overcome the challenge of FCL, we propose a novel approach named \ourapproach.

    Similar to typical FL approaches, the proposed \ourapproach operates through iterative communication rounds, during which participating clients cooperate with the server to learn a global model. Within each communication round, \ourapproach involves three primary steps to learn a global model: (1) \textbf{history-aware local inference}, (2) \textbf{local likelihood extraction} and (3) \textbf{global knowledge integration}. After one communication round, a global BNN model is learned, which integrates historical knowledge accumulated from both current and previous rounds. The posterior of the global BNN model after communication round $T$ can be denoted as:
	\begin{eqnarray}
		p(\theta | \distD^{1 \ldots T}) = p(\theta | \distD^{1 \ldots T-1} \cup \distD^T).
	\end{eqnarray}
	
	\textbf{History-aware local inference} At the start of communication round $T$, clients receive the latest global model of the last round: $ p(\theta | \distD^{1 \ldots T-1})$. Meanwhile, local data distribution on clients evolves into $D^{T}$. To integrate the new local distribution into the BNN posterior, we perform variational inference locally via Bayes by Backprop:
	\begin{eqnarray}
		p(\theta | \distD^{1 \ldots T-1} \cup \distD_k^T)
		\propto p(\theta | \distD^{1 \ldots T-1}) p( \distD_k^T | \theta).
		\label{eqn:local_vi}
	\end{eqnarray}
	After local inference, the local model adaptively learns from the latest local distribution and still retains knowledge about historical distributions.
	
	\textbf{Local likelihood extraction} Local inference directly optimizes the approximate posterior $p(\theta | \distD^{1 \ldots T-1} \cup \distD_k^T)$, and does not explicitly solve for the likelihood $p( \distD_k^T | \theta)$. In order to integrate separate local models into a global model, we extract local likelihood from local posterior. By rewriting Eq. (\ref{eqn:local_vi}),the following relation holds:
	\begin{eqnarray}
		p( \distD_k^T | \theta) \propto \frac{p(\theta | \distD^{1 \ldots T-1} \cup \distD_k^T)}{p(\theta | \distD^{1 \ldots T-1})}.
		\label{eqn:likelihood_inference}
	\end{eqnarray}
	Since the numerator and denominator are both Gaussian distributions, the quotient is also a Gaussian distribution with a constant factor
	\footnote{
		Specifically, the following identity holds:
		$\frac{\mathcal{N}(\mu_1,\Sigma_1)}{\mathcal{N}(\mu_2,\Sigma_2)} \propto \mathcal{N}(\mu,\Sigma)$,
		where $\Sigma = (\Sigma_1^{-1} -\Sigma_2^{-1})^{-1}, \mu = (\Sigma_1^{-1}\mu_1-\Sigma_2^{-1}\mu_2)\Sigma$ .
	}. Therefore, we can represent the local likelihood $p(D^T_k | \theta)$ by another Gaussian distribution.
	
	Intuitively, local likelihood encapsulates knowledge about the latest local data distribution. After extraction, local likelihoods on clients are sent to the server.
	
	\textbf{Global knowledge integration} With local likelihoods and posterior of the last round, a global model is aggregated on the server. Similar to Eq. (\ref{eqn:vcl_likelihood_decomp}) in VCL, the global distribution can be decomposed as:
	\begin{eqnarray}
		\begin{aligned}
			p(\theta | \distD^{1 \ldots T-1} \cup \distD^T)
			& \propto p(\theta | \distD^{1 \ldots T-1}) p( \distD^T | \theta) \\
			& \propto p(\theta | \distD^{1 \ldots T-1}) \prod_{1}^{k}{p( \distD_k^T | \theta)}.
			\label{eqn:fcl_likelihood_decomp}
		\end{aligned}
	\end{eqnarray}
	With Eq. (\ref{eqn:fcl_likelihood_decomp}), we can calculate the global posterior on the server, which is a product of multiple Gaussian densities and can be calculated in a closed form. The resulting global posterior $p(\theta | \distD^{1 \ldots T-1} \cup \distD^T)$ integrates knowledge of all the clients over all previous rounds and is used for training in the next round.

	Upon completion of a round of \ourapproach training, the latest local distributions separate on clients are integrated into the global BNN model, along with historical knowledge. Notably, since \ourapproach continuously integrates the latest data distribution during each round, no explicit task boundary is required, making \ourapproach applicable to all kinds of FCL scenarios.


\subsection{Prototype Library for Dynamic Label Space}
As a particular case of distribution change, new categories may emerge during the learning process. In a neural network, individual neurons in the output layer correspond to distinct categories in the label space. Thus, new categories cannot be handled by a model with static architecture. 
In some CL approaches\cite{classifier21Kuo, classifier22Chen}, the network is usually constructed by two modules: a \textit{feature extractor} and a \textit{classifier}. Currently, to handle new classes, many works assign a new classifier to the network when a new class emerges \cite{gem2017David, vcl18nguyen}.

We inherit the concept of feature extractor and classifier, but propose a distinct approach to handle the dynamic label space as shown in \cref{fig:prototype_lib}. In our case, we use the last full-connection layer as the classifier and the rest as the shared feature extractor (the shared model). In a sense, each neuron in the classifier can be regarded as a learned \textit{prototype} of the corresponding class. Essentially, a \textit{prototype library} is maintained by the clients, which is a table that stores the encountered labels and the corresponding prototypes. Before training starts, a classifier is assembled to build a full model by fetching all the prototypes corresponding to classes present in the training batch. After training on the full model, the classifier is split back into neurons and used to update the prototype library. For prediction, a classifier is built from the whole prototype library.
	The essential advantage of the prototype library is that it allows direct classification among all seen classes.  
	Moreover, an output neuron will only receive a negative loss and no positive loss if the batch has no samples of its class, which will harm the model's performance. Our approach mitigates this limitation by only assembling neurons associated with classes present in the batch.
	
	We also adopt the technique of warm-up used in continual learning. When new classes occur, an empty neuron is initialized for it. Classifier containing these neurons is considered \textit{mismatched} with the shared model. Essentially, these empty neurons induce significant loss to the shared model, making parameters in the shared model change dramatically. For the mismatched classifiers, a warm-up is performed before regular training by several epochs of fine-tuning on the batch, freezing the shared model.
	
	By aggregating entries from different tables that have the same label, a global prototype library is obtained. This helps ensemble client knowledge on the classification of the same category. Meanwhile, this also helps share class information with the clients who have yet to encounter a particular class. In addition, the global prototype library can be readily used for prediction.
	

	\subsection{SNN Based Initialization}
	
		\begin{figure}[t]
		\centering
		\begin{subfigure}[t]{\linewidth}
			\includegraphics[width=\linewidth]{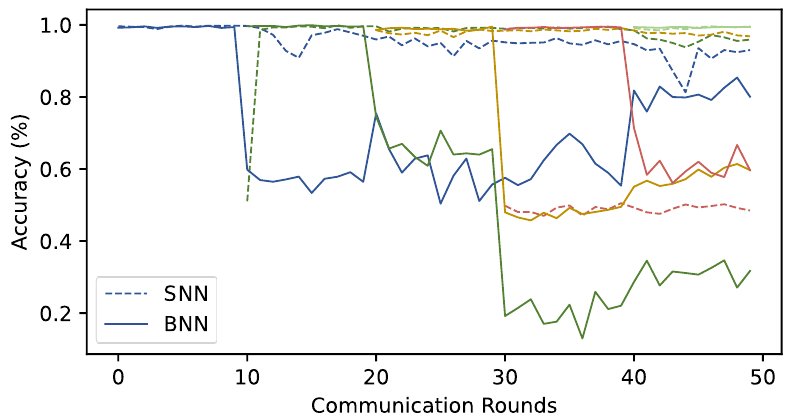}
			\caption{An example of class-incremental learning on MNIST. The dataset is divided into 5 splits containing number \{0,1\},\{2,3\},\{4,5\},\{6,7\},\{8,9\}.}
			\label{fig:snn_on_mnist}
		\end{subfigure}
		\begin{subfigure}[t]{\linewidth}
			\includegraphics[width=\linewidth]{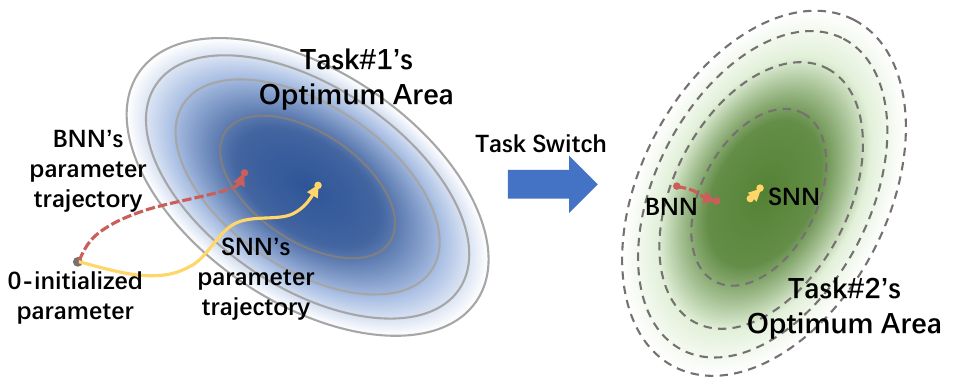}
			\caption{Demonstration of prior's effect in a simulated 2-D model solution space. Note that on task~1, BNN's parameter is drawn near the border. Consequently, when switched to task~2, the parameter change is more significant for BNN. }
			\label{fig:solution_space}
		\end{subfigure}
		\caption{On small datasets, the performance degradation of BNN is more significant than that of SNN after task switch, due to prior's effect of BNN.}
		\label{fig:bnn_prior}
	\end{figure}

	We found that on some small datasets, standard neural network (SNN) tends to exhibit no noticeable performance degradation, whereas BNN suffers from catastrophic forgetting (see \cref{fig:snn_on_mnist}). After in-depth analysis, we ascribe the difference to the effect of prior in BNN. Conventionally, the prior of BNN is set to a 0-centered normal distribution for the sake of simplicity, which suffices for stationary datasets. However, in the case of dynamic distribution, this leads to a problem, as depicted in \cref{fig:solution_space}. 
	
	Previous research on BNN reveals that the prior is a strong regularization for the model, drawing the posterior close to it. Since the 0-centered prior generally does not lie in the optimum area of the solution space, the BNN is optimized to the optimum area but remains proximal to the border near the prior. Conversely, unconstrained optimization on SNN typically culminates in convergence toward the center of the optimum area. When a slight change in the data distribution happens, the optimum area will drift, so networks are optimized to fit the new distribution. For SNN whose parameter is already located in the center of previous tasks' optimum area, only small steps are required to fit the new distribution. However, for BNN, whose parameter lies around the border, larger changes are taken to reach the optimum of changed distribution. Since the change in the parameter is a source of catastrophic forgetting, this explains the observed phenomenon.
	
	Critically, the initial 0-centered prior contains no actual information about data distribution, causing inferior performance than SNN. 
	To overcome this, we suggest the prior initialized from an unconstrained SNN learned from the data distribution. In the probabilistic view, SNN represents the max-a-posteriori (MAP) estimation of the parameter distribution. Consequently, the parameters learned by the SNN serve as a promising initialization for subsequent Bayesian Neural Network (BNN) training endeavors.
	
	Practically, in our algorithm, in the first several rounds, we train an SNN model using the classical FedAvg approach. Afterward, the server converts the global SNN model into an isomorphic BNN model. The mean of the BNN parameter is set to the corresponding SNN parameter value, and the variance is set to a predetermined value. This distribution is used as the prior and the initial value of the BNN model. In the following rounds, the BNN model is used for training as proposed.

	\subsection{Discussion and Limitations}
	
		\setlength{\textfloatsep}{1.2ex}
	\begin{algorithm}[t]
		\caption{The \ourapproach framework}
		\label{alg:ourapproach}
		\textbf{Input}: server, $K$ clients $c$, local streaming data on each client $d_k^r$. \\
		\textbf{Parameter}: SNN initialization rounds $r_{init}$, training parameters. \\
		\textbf{Output}: The global model and prototype library $\theta^r, H^r$.
		\begin{algorithmic}[1] 
			\STATE Initialize the global model $\theta^0$ as an SNN model, the prototype library $H^0$ as empty.
			\WHILE{the system running, round $r=1,2,\ldots$}
			\STATE Send the model $\theta^{r-1}$ and prototype library $H^{r-1}$ to clients.
			\FOR{ client $c_k$ in all clients}
			\STATE Assemble classifier $h_k^{r-1}$ from $H^{r-1}$.
			\STATE Local model $(\theta|h)_k^r \Leftarrow training((\theta|h)^{r-1},d_k^r)$ via SGD or Bayes by Backprop.
			\STATE Update prototype library $H_k^r$ from head $h_k^r$.
			\STATE Compute the local likelihood $p( \distD_k^T | \theta)$ using Eq.\ref{eqn:likelihood_inference} if model is BNN.
			\STATE Send local likelihood or local SNN model and prototype library $H_k^r$ to server.
			\ENDFOR
			\STATE Aggregate using aggregation of FedAvg for SNN or Eq.\ref{eqn:fcl_likelihood_decomp} for BNN.
			\STATE Set the prior to global posterior if model is BNN.
			\STATE \textbf{if} $r=r_{init}$ : Transform $\theta^r$,$H^r$ into BNN models.
			\ENDWHILE
		\end{algorithmic}
	\end{algorithm}

	Our approach \ourapproach is summarized in Alg. \ref{alg:ourapproach}. For simplicity, the algorithm assumes full participation for all clients. However, partial participation is allowed by performing local training on selected clients while broadcasting the new global model to all clients.
	In the paper, we focus on a global prototype library, which may leak local label information in highly privacy sensitive scenarios. In these cases, the prototype library can be kept locally, and only the feature extractor is shared. The local prototype library can also be utilized as a means of personalization\cite{to_pfl}. Furthermore, the use of BNN provides uncertainty information about prediction, which is crucial in scenarios like autonomous driving and medical diagnostics \cite{uncertainty_diag2023Duan, uncertaintycam2021Vakhitov,fluncertainty2023Zhu}. We leave these potential extensions for future research.
	


	\section{Experiments}

	\begin{table*}[t]
		\tiny
		\centering
		\resizebox{0.95\linewidth}{!}{
			\begin{tabular}{|c|c|c|c|c|c|c|c|c|c|c|}
				\hline \multicolumn{11}{|c|}{\textbf{Cifar-100 Class-Incremental}} \\ \hline
				\multirow{2}{*}{Methods} &  \multicolumn{2}{c|}{Class Group 1} & \multicolumn{2}{c|}{Class Group 2} & \multicolumn{2}{c|}{Class Group 3} & \multicolumn{2}{c|}{Class Group 4} & \multicolumn{2}{c|}{Average} \\ \cline{2-11}
				& @TS & @Fin & @TS & @Fin & @TS & @Fin & @TS & @Fin & @TS & @Fin \\ \hline
				FedAvg & 18.86 & 4.58 & 25.11 & 8.93 & 36.94 & 8.93 & 38.50 & 38.50 & 29.85 & 15.23 \\
				SCAFFOLD & 26.12 & 17.30 & 23.77 & 21.43 & 38.84 & 28.91 & 28.12 & 28.12 & 29.21 & 23.94 \\
				EWC + FL & 24.91 & 5.25 & 17.48 & 6.81 & 5.47 & 7.25 & 7.70 & 7.70 & 13.89 & 6.75 \\
				MAS + FL & 15.96 & 16.07 & 12.39 & 13.39 & 19.42 & 18.08 & 20.20 & 20.20 & 16.99 & 16.94 \\
				LwF + FL & 29.35 & 5.80 & \textbf{31.47} & 10.27 & \textbf{42.75} & 18.42 & \textbf{36.38} & \textbf{36.38} & \textbf{34.99} & 17.72 \\
				FCIL & 30.89 & 22.18 & 25.27 & 18.51 & 36.45 & 24.12 & 33.88 & 33.88 & 31.62 & 24.67 \\ \hline
				\textbf{\ourapproach} & \textbf{33.37} & \textbf{28.73} & 25.40 & \textbf{21.67} & 30.95 & \textbf{29.94} & 30.04 & 30.04 & 29.94 & \textbf{27.60} \\
				\hline \hline 
				\multicolumn{11}{|c|}{\textbf{Tiny-ImageNet Class-Incremental}} \\ \hline
				\multirow{2}{*}{Methods} &  \multicolumn{2}{c|}{Class Group 1} & \multicolumn{2}{c|}{Class Group 2} & \multicolumn{2}{c|}{Class Group 3} & \multicolumn{2}{c|}{Class Group 4} & \multicolumn{2}{c|}{Average} \\ \cline{2-11}
				& @TS & @Fin & @TS & @Fin & @TS & @Fin & @TS & @Fin & @TS & @Fin \\ \hline
				FedAvg & 9.01 & 5.79 & 5.90 & 2.34 & 2.12 & 2.79 & 2.01 & 2.01 & 4.76 & 3.23 \\
				SCAFFOLD & \textbf{12.72} & 7.48 & 13.62 & 9.15 & 13.84 & 10.94 & 16.18 & 16.18 & 14.09 & 10.94\\
				EWC + FL & 4.24 & 4.02 & 4.69 & 5.25 & 9.15 & 8.04 & 8.48 & 8.48 & 6.64 & 6.45 \\
				MAS + FL & 7.90 & 5.34 & 6.01 & 2.12 & 2.34 & 2.79 & 2.12 & 2.12 & 4.59 & 3.09 \\
				LwF + FL & 10.95 & 2.46 & \textbf{17.00} & 6.03 & 15.10 & 13.75 & 15.88 & 15.88 & 14.73 & 9.53 \\
				FCIL & 10.69 & \textbf{7.51} & 14.90 & 10.37 & 14.49 & \textbf{13.84} & 14.34 & 14.34 & 13.61 & 11.50 \\ \hline
				\textbf{\ourapproach} & 10.48 & 7.31 & 13.52 & \textbf{10.41} & \textbf{16.17} & 13.75 & \textbf{19.93} & \textbf{19.93} & \textbf{15.02} & \textbf{12.85} \\ \hline
			\end{tabular}
		}
		\caption{The test accuracy (\%) of each task at the round of task switch (@TS), and after the final round(@Fin), of the class-incremental settings.}
		\label{tab:ci_res}
		\vspace{-1.0em}
	\end{table*}
	
	\begin{table*}[t]
		\tiny
		\centering
		\resizebox{0.95\linewidth}{!}{
			\begin{tabular}{|c|c|c|c|c|c||c|c|c|c|}
				\hline
				\multicolumn{2}{|c|}{} & \multicolumn{4}{c||}{\textbf{Small-Scale Image Classification}} &  \multicolumn{4}{c|}{\textbf{Large-Scale Image Classification}}\\ \hline
				\multicolumn{2}{|c|}{Methods} &  Task 1 & Task 2 & Task 3 &  Avg. &  Task 1 & Task 2 & Task 3 &  Avg. \\ \hline
				\multirow{2}{*}{FedAvg} & @TS & 10.94 & 13.84 & 5.25 & 10.01 &29.13 & 10.88 & 5.13 & 15.04 \\ \cline{2-10}
				& @Fin & 10.60 & 4.24 & 5.25 & 6.70 & 9.25 & 3.98 & 5.13 & 6.12 \\ \hline
				\multirow{2}{*}{SCAFFOLD} & @TS & 26.49 & 4.58 & 4.00 & 11.69 & \textbf{37.50} & 8.25 & 6.00 & 17.25 \\ \cline{2-10}
				& @Fin & 21.92 & 3.68 & 4.00 & 9.87  & 27.38 & 7.75 & 6.00 & 13.71 \\ \hline
				\multirow{2}{*}{EWC+FL} & @TS& 25.40 & 2.23 & 3.78 & 10.47 & 32.00 & 12.75 & 4.75 & 16.50 \\ \cline{2-10}
				& @Fin & \textbf{22.54} & 2.23 & 3.78 & 9.52 & 16.50 & 6.12 & 4.75 & 9.13 \\ \hline
				\multirow{2}{*}{MAS+FL} & @TS & 11.16 & 11.45 & 4.56 & 9.06 & 12.75 & 10.88 & 3.50 & 9.04\\ \cline{2-10}
				& @Fin & 10.60 & 5.89 & 4.56 & 7.02 & 11.50 & 4.00 & 3.50 & 6.33 \\ \hline
				\multirow{2}{*}{LwF+FL} & @TS & 34.17 & \textbf{15.78} & 8.26 & 19.40 & 33.50 & \textbf{17.87} & \textbf{6.13} & 19.17 \\ \cline{2-10}
				& @Fin & 11.27 & 5.25 & 8.26 & 8.26 & 11.87 & 4.87 & \textbf{6.13} & 7.62 \\ \hline
				\multirow{2}{*}{\textbf{\ourapproach}} & @TS & \textbf{44.45} & 13.37 & \textbf{8.45} & \textbf{22.09} & 37.28 & 14.58 & 5.91 & \textbf{19.25} \\ \cline{2-10}
				& @Fin & 22.23 & \textbf{13.09} & \textbf{8.45} & \textbf{14.59} & \textbf{31.83} & \textbf{12.80} & 5.91 & \textbf{16.85} \\ \hline
			\end{tabular}
		}
		\caption{The test accuracy (\%) of each task at the round of task switch (@TS), and after the final round(@Fin), of the task-incremental settings.}
		\label{tab:ti_res}
		\vspace{-1.5em}
	\end{table*}

	\subsection{Experimental Setup}
	There are currently few approaches designed specifically for FCL problems. Therefore, following existing works on FCL, we implement some continual learning approaches on top of federated learning frameworks:
    1) \textbf{Learning without Forgetting (LwF + FL)}, which performs local distillation following the proposal of LwF \cite{LwF2016Li};
    2) \textbf{Elastic Weight Consolidation (EWC + FL)}, which regularizes local update with EWC \cite{ewc2017kirkpatrick};
    3) \textbf{Memory Aware Synapse (MAS + FL)}, which regularizes local update with MAS \cite{mas2018Aljundi}. 
    We also include some federated learning algorithms in the comparison: the vanilla \textbf{FedAvg} \cite{fl2017mcmahan} and \textbf{SCAFFOLD} \cite{scaffold2020Karimireddy}, a state-of-art algorithm for heterogeneous federated learning. Since these algorithms cannot deal with dynamic label space by nature, we implement them with the same prototype library as ours. 
	We also include the federated class-incremental learning approach \textbf{FCIL} \cite{fcil2022Dong} in the experiments on federated class-incremental settings.

	\begin{table*}[t]
        \tiny
		\centering
		\resizebox{0.95\linewidth}{!}{
			\begin{tabular}{|c|c|c|c|c|c||c|c|c|c|}
				\hline
				& \multicolumn{5}{c||}{\textbf{Class-Incr, Tiny-ImageNet}} &  \multicolumn{4}{c|}{\textbf{Task-Incr, Image Classification}}\\ \hline
				Methods &  Task 1 & Task 2 & Task 3 & Task 4 &  Avg. &  Task 1 & Task 2 & Task 3 &  Avg. \\ \hline
				FedAvg & 2.12 & 2.01 & 2.34 & 2.57 & 2.26 & 12.05 & 3.01 & 6.70 & 7.25 \\ \hline
				SCAFFOLD & 8.04 & 11.94 & 12.39 & 19.42 & 12.95 & 37.21 & 5.02 & 1.56 & 14.60 \\ \hline
				EWC+FL & 5.47 & 7.59 & 7.92 & 12.72 & 8.43 & 17.86 & 2.34 & 1.12 & 7.11 \\ \hline
				MAS+FL & 2.34 & 2.23 & 2.34 & 2.57 & 2.37 & 11.05 & 1.12 & 0.89 & 4.35 \\ \hline
				LwF+FL & 6.03 & 11.83 & \textbf{24.33} & 27.23 & 17.35 & 15.96 & 8.48 & \textbf{8.93} & 11.12 \\ \hline
				\textbf{\ourapproach} & \textbf{8.04} & \textbf{18.28} & 21.45 & \textbf{30.34} & \textbf{19.53} & \textbf{42.56} & \textbf{15.23} & 8.29 & \textbf{22.03} \\ \hline
			\end{tabular}
		}
		\caption{The final accuracy (\%) under gradual FCL settings.}
		\label{tab:grad_res}
        \vspace{-1.0em}
	\end{table*}

	The hyperparameters in the experiments are set according to the original proposal or by tuning on the validation set. For a fair comparison, \ourapproach uses Bayesian neural network models, and the others use standard neural networks with the same architecture as the BNN. All the experiments are run with 100 participating clients, and 10 of them are selected to participate in each round of training, unless stated otherwise. Detailed experiment settings and further experiment results (standard deviations, experiments with varying participation rates, etc.) are included in the supplementary material.
	
	

	\subsection{Task-Separate Federated Continual Settings}
	We first evaluate the performance on the classical task-separate federated continual settings, where all the clients change to some new task simultaneously, while the local distributions across clients are still non-IID (corresponding to \cref{fig:fcl_case_1}). Specifically, we tested the algorithms with two class-incremental and two task-incremental settings:
	1) Class-incremental learning on Cifar-100 \cite{cifar09Alex}, split into 4 tasks, each with 25 classes. 2) Class-incremental learning on Tiny-ImageNet \cite{timnet15Olga}, split into 4 tasks, each with 50 classes. 3) {Task-Incremental classification on small scale images}, using Cifar-10 $\rightarrow$ Cifar-100 $\rightarrow$ Tiny-ImageNet. \cite{usps94Hull,mnist98Lecun,emnist17Cohen}. 4) {Task-Incremental classification on larger scale images}, using using STL-10 $\rightarrow$ Flowers-102 $\rightarrow$ Food-101 \cite{stl11coates,fl10208Nilsback,food10114bossard}.
	
	The accuracy of the compared algorithms is evaluated both at the round of task switch (@TS) and the final round (@Fin), as detailed in \cref{tab:ci_res} and \cref{tab:ti_res}. It can be observed that the performance of compared baselines varies significantly. While some algorithms (FedAvg, SCAFFOLD) suffer from significant forgetting, some (EWC + FL, MAS + FL) maintains their performance on previous tasks but fail to learn subsequent tasks. This reflects the well-known plasticity-stability dilemma of continual learning \cite{spdilemma05Abraham, spdilemma23Jung}. Among the compared baselines, \ourapproach demonstrates superior ability in mitigating forgetting while exhibiting comparable adaptability in learning new tasks, consistently reaching the highest average accuracy. Notably, the mere adaptation of continual learning approaches to federated learning settings falls short in effectively addressing the challenges posed by FCL. This highlights the necessity of developing specifically designed algorithms for FCL scenarios.

	\subsection{Gradual Distribution Change}
	We explore the proposed approach in a more realistic setting, where the task change happens gradually. Specifically, we evaluate the approaches in a class-incremental setting and a task-incremental setting:
	1) Gradual federated class-incremental setting, using Tiny-ImageNet, where new classes emerge gradually on clients. 2) Gradual federated task-incremental setting of image classification, using task sequence of Cifar-10 $\rightarrow$ Cifar-100 $\rightarrow$ Tiny-ImageNet.
	
	\begin{figure}[t]
		\centering
		\begin{subfigure}[t]{\linewidth}
		      \centering
			\includegraphics[width=0.85\linewidth]{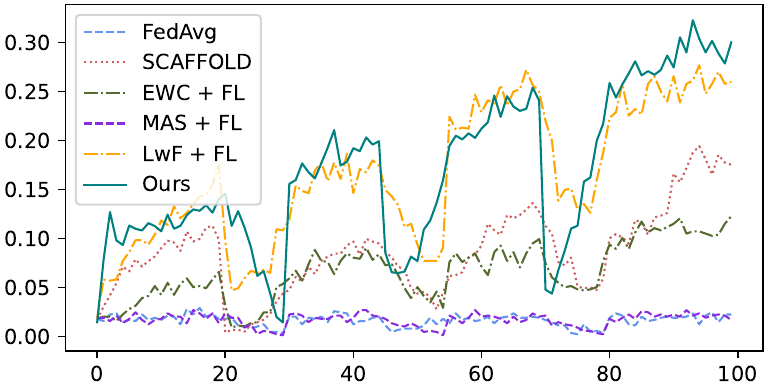}
			\caption{Gradual class-incremental on Tiny-ImageNet}
			\label{fig:exp_grad_ci}
		\end{subfigure}
		\begin{subfigure}[t]{\linewidth}
            \centering
			\includegraphics[width=0.85\linewidth]{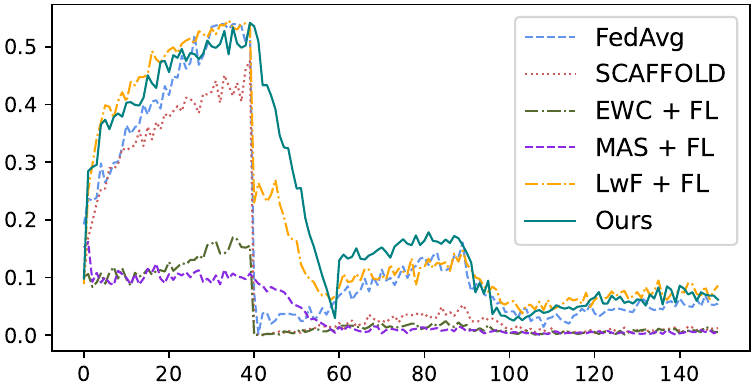}
			\caption{Gradual task-incremental on image-classification.}
			\label{fig:exp_grad_ti}
		\end{subfigure}
		\caption{Focus-On-Now (FON) accuracy curves of gradual distribution change settings, which denotes the model accuracy on current data distribution.}
		\label{fig:exp_gradual_fon_curve}
	\end{figure}
	
	\cref{tab:grad_res} reports the final accuracy of compared algorithms. Similar to results in task-separated settings, \ourapproach performs better than the other baselines.
	   Since there are no explicit task boundaries in gradual FCL scenarios, we also propose to measure the performance of algorithms by Focus-on-Now (FON) accuracy curves, as shown in \cref{fig:exp_gradual_fon_curve}. As the name suggests, the FON curve records model accuracy on \textit{current} global data distribution, which reflects the model performance the user can experience \textit{now} during the system running. It can be found that curves for \ourapproach usually lie above the other baselines, especially during the phase of distribution change.  This means that the model for \ourapproach adapts well to the current distribution and provides better inference performance for federated continual learning tasks, which is a desirable attribute for real world FCL applications.
	
	\section{Conclusion}
	In this paper, we first formulate the problem of Federated Continual Learning, which takes the temporal dynamic of data distribution into consideration beyond classical federated learning. Then, we propose \ourapproach to tackle the general FCL challenge. \ourapproach utilizes the Bayesian neural network, and integrates knowledge about historical and local distribution into one global model. Through extensive experiments, \ourapproach is shown to outperform FL and FCL baselines in various settings.

\bibliographystyle{IEEEtran}
\bibliography{reference}

\clearpage

\appendices

\section{Detailed Algorithm}
In this section, we present a more detailed description of the proposed \ourapproach algorithm (\cref{alg:detailed}). The loop continues until the system stops running, but at any time, the latest head library and shared model can be used to perform inference on the server or clients. For the sake of simplicity, the algorithm assumes full participation for all clients. However, partial participation is allowed by performing local training on selected clients, while broadcasting new global model to all clients.

\begin{algorithm}[htbp]
	\caption{\ourapproach}
	\label{alg:detailed}
	\textbf{Input}: server, $K$ clients $c$, local streaming data on each client $d_k^r$. \\
	\textbf{Parameter}: SNN initialization rounds $r_{init}$, training parameters. \\
	\textbf{Output}: The global model and prototype library for prediction $\theta^r$,$H^r$
	\begin{algorithmic}[1] 
		\STATE Initialize the global model $\theta^0$ as an SNN model, the prototype library $H^0$ as empty.
		\WHILE{the system running, round $r=1,2,\ldots$}
		\STATE Send the model $\theta^{r-1}$ and prototype library $H^{r-1}$ to clients.
		\FOR{ client $c_k$ in all clients}
		\STATE Client training using \cref{alg:client_training}.
		\ENDFOR
		\STATE Server aggregation using \cref{alg:server_aggregation}.
		\IF {$r=r_{init}$}
		\STATE Transform $\theta^r$,$H^r$ into BNN models.
		\ENDIF
		\ENDWHILE
	\end{algorithmic}
\end{algorithm}

\begin{algorithm}[htbp]
	\caption{Client training}
	\label{alg:client_training}
	\textbf{Input}: Global model and prototype library of last round $\theta^{r-1}$, $H^{r-1}$
	\begin{algorithmic}[1] 
		\STATE Assemble classifier $h_k^{r-1}$ from $H^{r-1}$.
		\IF {local model is SNN}
		\STATE Local model $(\theta|h)_k^r \Leftarrow training((\theta|h)^{r-1},d_k^r)$ via SGD.
		\STATE Update prototype library $H_k^r$ from head $h_k^r$.
		\STATE Send local model $\theta_k^r$ and prototype library $H_k^r$ to server.
		\ELSE
		\STATE Local model $(\theta|h)_k^r \Leftarrow training((\theta|h)^{r-1},d_k^r)$ via Bayes by Backprop.
		\STATE Update prototype library $H_k^r$ from head $h_k^r$.
		\STATE Calculate the likelihood by Eq.\ref{eqn:likelihood_inference}.
		\STATE Send local likelihood $p(D^T_k | \theta)$ and prototype library $H_k^r$ to server.
		\ENDIF
	\end{algorithmic}
\end{algorithm}

\begin{algorithm}[htbp]
	\caption{Server aggregation}
	\label{alg:server_aggregation}
	\textbf{Input}: local prototype libraries $H_k^r$, local models $\theta_k^r$ or local likelihood $\{p(D^T_k | \theta)\}$ and latest global model $\theta^{r-1}$. \\
	\textbf{Output}: The aggregated global model and prototype library  $\theta^r$,$H^r$
	\begin{algorithmic}[1] 
		\STATE Aggregate the prototype library $H^r \Leftarrow agg(\{H_k^r\})$.
		\IF {local models are SNN}
		\STATE Aggregate using aggregation of FedAvg : $\theta^r \Leftarrow agg(\{\theta_k^r\})$.
		\ELSE
		\STATE Aggregate using Equation \ref{eqn:fcl_likelihood_decomp} : $\theta^r \Leftarrow agg(\theta^{r-1},\{p(D^T_k | \theta)\})$.
		\ENDIF
	\end{algorithmic}
\end{algorithm}

\section{Detailed Experimental Setup}
In this section, we elaborate the details of experiments, including datasets and hyper-parameters. All the experiments are deployed on a x86\_64 server with 2x Intel 40-core CPUs, 160GB memory, and 2x NVIDIA RTX3090 GPUs.

\subsection{Datasets}
\textbf{Cifar-10} \cite{cifar09Alex} is a popular image classification benchmark dataset, which consists of 60,000 colored images with the resolution of 32x32, covering 10 classes, with 6,000 images per class. The dataset is divided into a training set of 50,000 samples and a test set of 10,000 samples.

\textbf{Cifar-100} \cite{cifar09Alex} is an extension of Cifar-10, which contains 100 classes. Each class contains 500 training samples and 100 test samples.

\textbf{Tiny-ImageNet} is an image classification dataset extracted from ImageNet dataset \cite{timnet15Olga}. It contains 200 selected classes, each with 500 training images, 50 validating images and 50 test images. All the samples are 64x64 colored images. In the class-incremental experiment, the original images are used. In the task-incremental experiment, the samples are resized to 32x32 colored images to match the dimension of Cifar-10 and Cifar-100.

\textbf{STL-10} \cite{stl11coates} is an image recognition dataset inspired by Cifar-10 dataset with some improvements. The dataset contains 10 classes with 500 training samples and 800 test samples per class. The samples are 96x96 pixel colored images.

\textbf{Flowers-102} (Oxford 102 flowers) \cite{fl10208Nilsback} dataset is a consistent of 102 flower categories commonly occurring in the United Kingdom. The training and test set consist of 6,149 and  1020 colored images respectively. The images have large scale, pose and light variations. In addition, there are categories that have large variations within the category and several very similar categories. In the task-incremental experiment, the samples are resized to 96x96 images to match the dimension with other tasks.

\textbf{Food-101} \cite{food10114bossard} is an image classification dataset consisting of 101 food categories. For each class, 750 training samples are provided as well as 250 test samples. On purpose, the training images were not cleaned, and thus still contain some amount of noise. In the task-incremental experiment, the samples are resized to 96x96 colored images to match the dimension with other tasks.

\subsection{Data Generation Strategy}
To simulate the complicated data evolution patterns of the various FCL cases discussed in the paper, we implement an FCL data generation simulator for the experiments. Here, we describe each of the data generation strategies used in the different experiment settings.

\textbf{Task Separate FCL Setting} In the task separate class-incremental and task-incremental setting, class groups and datasets are used as tasks respectively. Each task spans several rounds, while switch to the next task happens at certain rounds, known as task boundaries. The task samples are used as the global data during the rounds of each task. To simulated the non-IID characteristic of client data, global data are partitioned based on Dirichlet distribution into client datasets, as described in \cite{dirichlet19Hsu}. The task boundaries of each setting tested in our experiment is listed in \cref{task_boundary}.

\begin{table*} [ht]
	\centering 
	\caption{The task boundaries of tested settings in the experiment.}
	\label{task_boundary}
	\begin{tabular}{c|c} \hline
		Experiment Setting & Task Boundaries (Round\#) \\ \hline
		Cifar-100 Class-Incremental & 4 tasks each with 25 classes: 25,50,75,100 \\ \hline
		Tiny-ImageNet Class-Incremental & 4 tasks each with 50 classes: 25,50,75,100 \\ \hline
		Small-Scale Object Classification & 3 tasks: 50,100,150 \\ \hline
		Larger-Scale Object Classification & 3 tasks: 50,100,150 \\ \hline
	\end{tabular}
\end{table*}

\begin{figure}
	\centering
	\includegraphics[width=\linewidth]{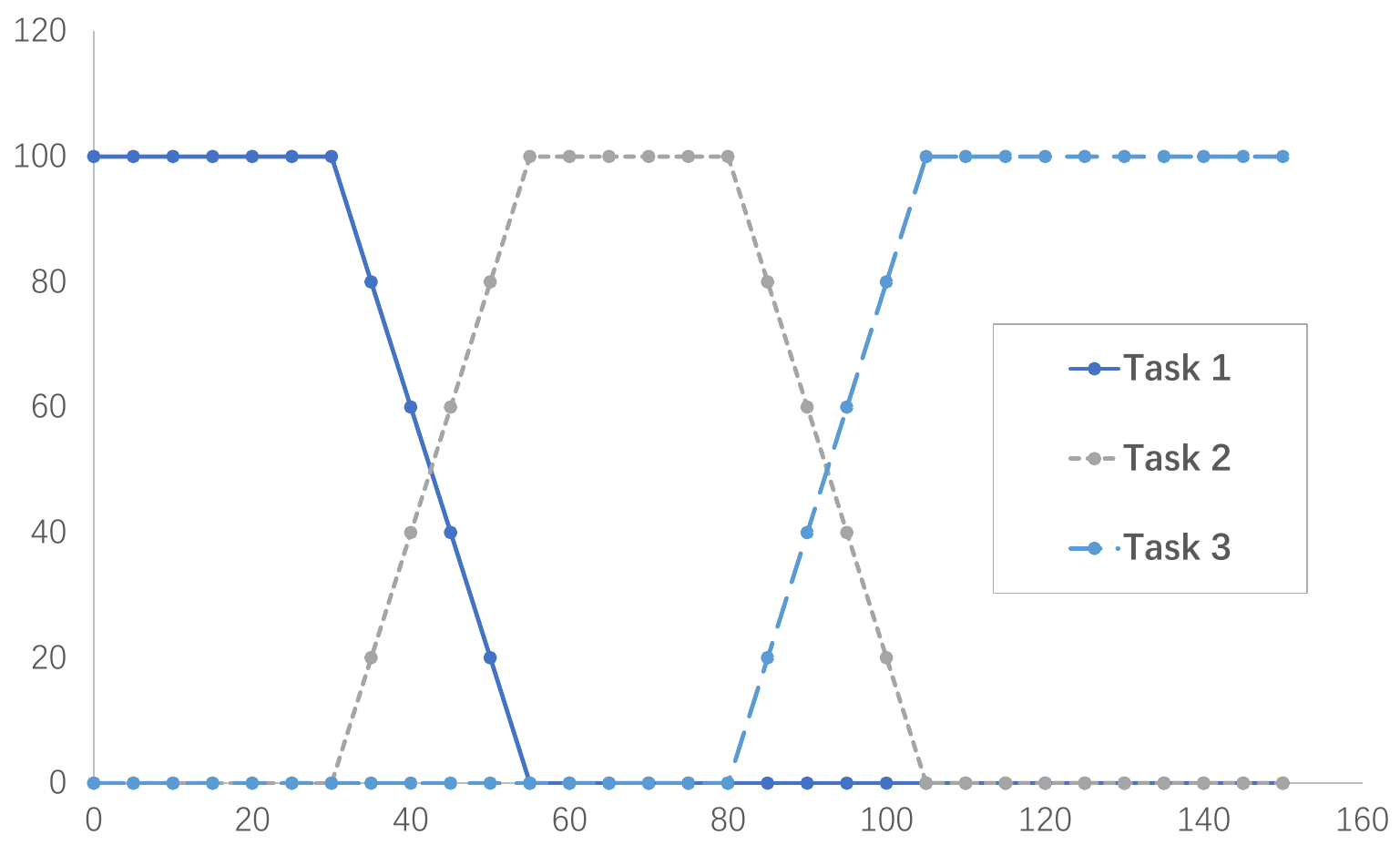}	
	\caption{Task switch pattern in the gradual FCL Setting. The vertical axis is the data proportion of each task, measured in percentage; the horizontal axis is the communication rounds.}
	\label{gradual_proportion}
\end{figure}

\textbf{Gradual FCL Setting} In this setting, each task is assigned with a period of rounds as their duration. The duration of each task overlaps with previous and following tasks. During overlapped rounds, global data is formed by samples from the two tasks. The proportion of the previous task will gradually decrease while samples from the next task will increase, as depicted in \cref{gradual_proportion}. To simulate the behavior that some clients change to the new task first, we first decide the number of clients for each task, corresponding to the task proportion. Then the task data is partitioned based on Dirichlet non-IID among the clients. The duration of each tasks in our experiment is listed in \cref{task_duration}.

\begin{table*} [ht]
	\centering 
	\caption{The task duration of tested gradual FCL settings in the experiment.}
	\label{task_duration}
	\begin{tabular}{c|c|c} \hline
		Setting & Task & Task Duration (Rounds\#) \\ \hline
		\multirow{4}{*}{Tiny-ImageNet Gradual Class-Incremental} & Class Group 1 & 0-30 \\ \cline{2-3}
		& Class Group 2 & 20-55 \\ \cline{2-3}
		& Class Group 3 & 45-80 \\ \cline{2-3}
		& Class Group 4 & 70-100 \\ \hline
		\multirow{3}{*}{Obj-Classification Gradual Task-Incremental} & Cifar-10 & 0 - 60 \\ \cline{2-3}
		& Cifar-100 & 40 - 110 \\ \cline{2-3}
		& Tiny-ImageNet & 90 - 150 \\ \hline
	\end{tabular}
\end{table*}

\subsection{Implementation Detail}

\textbf{Algorithm Implementation} In the experiment, we implemented \ourapproach and the baselines FedAvg, SCAFFOLD, EWC + FL, MAS + FL, and LwF + FL. All the implements are based on the PyTorch 1.12 machine learning framework\footnote{Project homepage: \url{https://pytorch.org/}}. For \ourapproach, we choose torchbnn\footnote{Project repository: \url{https://github.com/Harry24k/bayesian-neural-network-pytorch}} as the Bayesian neural network library. All the baselines are carefully implemented based on the original paper and reference source codes provided by the authors.

For the FCIL baseline \cite{fcil2022Dong}, we used the source code provided by the author\footnote{Project repository: \url{https://github.com/conditionWang/FCIL}}. We made proper modification to the source code regarding data generation and evaluation to match with other compared algorithms.

\textbf{Model Architecture}
In the paper, we use ResNet-18 for all the experiments and all the baselines. The detailed network architecture is specified in \cref{ResNet_arch}, while the structure of residual layer is demonstrated in \cref{ResLayer_arch}.

\begin{table*} [htbp]
	\renewcommand\arraystretch{2.5}
	\centering 
	\caption{Architecture of ResNet-18 model used in our experiment.}
	\label{ResNet_arch}
	\begin{tabular}{c|c} \hline
		Layer name & Description \\ \hline
		conv1 & \makecell{Conv2d layer, $3\to64$ channels, $7\times7$ kernel, stride 4, padding 1, \\ with BatchNorm, ReLU and $3\times3$ MaxPooling (matches input dimension)} \\
		res1 & \makecell{Residual layer, \\ $64\to64$ channels, stride 1 } \\
		res2 & \makecell{Residual layer, \\ $64\to128$ channels, stride 2 } \\
		res3 & \makecell{Residual layer, \\ $128\to256$ channels, stride 2 } \\
		res4 & \makecell{Residual layer, \\ $256\to512$ channels, stride 2 } \\
		fc1 & \makecell{Linear layer, $[ n_{res4\_out}, 1000 ]$ with ReLU \\ (Matches res4 output dimension)} \\
		classifier & \makecell{Linear layer, $[ 1000, n_{class} ]$ with SoftMax \\ (Matches output classes)} \\ \hline
	\end{tabular}
\end{table*}

\begin{table*} [htbp]
	\renewcommand\arraystretch{3}
	\centering 
	\caption{Architecture of residual layer in the ResNet-18 model.}
	\label{ResLayer_arch}
	\begin{tabular}{c|c} \hline
		Layer name & Description \\ \hline
		conv1 & \makecell{Conv2d layer, $n_{in}\to n_{out}$ channels, $3\times3$ kernel, stride $n_{stride}$, \\ padding 1, no bias, with BatchNorm and ReLU.} \\
		conv2 & \makecell{Conv2d layer, $n_{out}\to n_{out}$ channels, $3\times3$ kernel, stride 1, \\ padding 1, no bias, with BatchNorm.} \\
		downsample & \makecell{ Conv2d layer, present if $n_{stride}\neq 1$ or $n_{in} \neq n_{out}$, \\ $n_{in}\to n_{out}$ channels, $3\times3$ kernel, stride $n_{stride}$, \\ padding 1, no bias, with BatchNorm.} \\ \hline
		Computation & $output = conv2(conv1(input)) + downsample(input)$ \\ \hline
	\end{tabular}
\end{table*}

\section{Further Experiment Results}
In this section, we further present some empirical results of \ourapproach and the compared baselines. The experiment settings are consistent with experiments in the main paper and the previous section.

\subsection{Task-Separate Federated Continual Settings}
Due to page limitations, the standard deviation of numerical results are not reported in the main paper. In this section, we present the standard deviations of the results under task-separate FCL settings. \cref{tab:ci_res_dev} corresponds to \cref{tab:ci_res} in the main paper, while \cref{tab:ti_res_dev} corresponds to \cref{tab:ti_res}.

\subsection{Gradual Distribution Change}
Due to page limitations, the standard deviation of numerical results are not reported in the main paper. In this section, we present the standard deviations of the results under gradual FCL settings. \cref{tab:grad_res_dev} corresponds to \cref{tab:grad_res} in the main paper.

\begin{sidewaystable*}
	\centering
	\resizebox{1.\linewidth}{!}{
		\begin{tabular}{|c|c|c|c|c|c|c|c|c|c|c|}
			\hline \multicolumn{11}{|c|}{\textbf{Cifar-100 Class-Incremental}} \\ \hline
			\multirow{2}{*}{Methods} &  \multicolumn{2}{c|}{Class Group 1} & \multicolumn{2}{c|}{Class Group 2} & \multicolumn{2}{c|}{Class Group 3} & \multicolumn{2}{c|}{Class Group 4} & \multicolumn{2}{c|}{Average} \\ \cline{2-11}
			& @TS & @Fin & @TS & @Fin & @TS & @Fin & @TS & @Fin & @TS & @Fin \\ \hline
			FedAvg & 18.86 $\pm$ \small{0.90} & 4.58 $\pm$ \small{0.17} & 25.11 $\pm$ \small{0.06} & 8.93 $\pm$ \small{0.23} & 36.94 $\pm$ \small{0.94} & 8.93 $\pm$ \small{0.13} & 38.50 $\pm$ \small{1.81} & 38.50 $\pm$ \small{1.81} & 29.85 $\pm$ \small{0.43} & 15.23 $\pm$ \small{0.40} \\
			SCAFFOLD & 26.12 $\pm$ \small{1.05} & 17.30 $\pm$ \small{0.85} & 23.77 $\pm$ \small{1.71} & 21.43 $\pm$ \small{0.88} & 38.84 $\pm$ \small{2.11} & 28.91 $\pm$ \small{1.24} & 28.12 $\pm$ \small{0.86} & 28.12 $\pm$ \small{0.86} & 29.21 $\pm$ \small{0.59} & 23.94 $\pm$ \small{0.20} \\
			EWC + FL & 24.91 $\pm$ \small{0.52} & 5.25 $\pm$ \small{0.04} & 17.48 $\pm$ \small{0.63} & 6.81 $\pm$ \small{0.56} & 5.47 $\pm$ \small{0.29} & 7.25 $\pm$ \small{0.25} & 7.70 $\pm$ \small{0.01} & 7.70 $\pm$ \small{0.01} & 13.89 $\pm$ \small{0.35} & 6.75 $\pm$ \small{0.10} \\
			MAS + FL & 15.96 $\pm$ \small{0.98} & 16.07 $\pm$ \small{0.76} & 12.39 $\pm$ \small{0.66} & 13.39 $\pm$ \small{0.88} & 19.42 $\pm$ \small{0.60} & 18.08 $\pm$ \small{1.25} & 20.20 $\pm$ \small{0.49} & 20.20 $\pm$ \small{0.49} & 16.99 $\pm$ \small{0.28} & 16.93 $\pm$ \small{0.23} \\
			LwF + FL & 29.35 $\pm$ \small{2.27} & 5.80 $\pm$ \small{0.18} & \textbf{31.47} $\pm$ \small{0.31} & 10.27 $\pm$ \small{1.29} & \textbf{42.75} $\pm$ \small{0.36} & 18.42 $\pm$ \small{0.56} & \textbf{36.38} $\pm$ \small{0.53} & \textbf{36.38} $\pm$ \small{0.53} & \textbf{34.99} $\pm$ \small{0.52} & 17.72 $\pm$ \small{0.30} \\
			FCIL & 30.89 $\pm$ \small{0.14} & 22.18 $\pm$ \small{0.82} & 25.27 $\pm$ \small{1.00} & 18.51 $\pm$ \small{2.01} & 36.45 $\pm$ \small{1.14} & 24.12 $\pm$ \small{0.68} & 33.88 $\pm$ \small{0.62} & 33.88 $\pm$ \small{0.62} & 31.62 $\pm$ \small{0.40} & 24.67 $\pm$ \small{1.00} \\ \hline
			\textbf{\ourapproach} & \textbf{33.37} $\pm$ \small{0.59} & \textbf{28.73} $\pm$ \small{0.37} & 25.40 $\pm$ \small{0.91} & \textbf{21.67} $\pm$ \small{1.27} & 30.95 $\pm$ \small{0.85} & \textbf{29.94} $\pm$ \small{1.31} & 30.04 $\pm$ \small{0.75} & 30.04 $\pm$ \small{0.75} & 29.94 $\pm$ \small{0.32} & \textbf{27.59} $\pm$ \small{0.60} \\
			\hline \hline 
			\multicolumn{11}{|c|}{\textbf{Tiny-ImageNet Class-Incremental}} \\ \hline
			\multirow{2}{*}{Methods} &  \multicolumn{2}{c|}{Class Group 1} & \multicolumn{2}{c|}{Class Group 2} & \multicolumn{2}{c|}{Class Group 3} & \multicolumn{2}{c|}{Class Group 4} & \multicolumn{2}{c|}{Average} \\ \cline{2-11}
			& @TS & @Fin & @TS & @Fin & @TS & @Fin & @TS & @Fin & @TS & @Fin \\ \hline
			FedAvg & 9.01 $\pm$ \small{0.23} & 5.79 $\pm$ \small{0.20} & 5.90 $\pm$ \small{0.24} & 2.34 $\pm$ \small{0.09} & 2.12 $\pm$ \small{0.11} & 2.79 $\pm$ \small{0.09} & 2.01 $\pm$ \small{0.34} & 2.01 $\pm$ \small{0.34} & 4.76 $\pm$ \small{0.11} & 3.23 $\pm$ \small{0.14} \\
			SCAFFOLD & \textbf{12.72} $\pm$ \small{1.61} & 7.48 $\pm$ \small{0.24} & 13.62 $\pm$ \small{0.57} & 9.15 $\pm$ \small{0.26} & 13.84 $\pm$ \small{0.34} & 10.94 $\pm$ \small{0.47} & 16.18 $\pm$ \small{1.26} & 16.18 $\pm$ \small{1.26} & 14.09 $\pm$ \small{0.44} & 10.94 $\pm$ \small{0.16} \\
			EWC + FL & 4.24 $\pm$ \small{0.07} & 4.02 $\pm$ \small{0.29} & 4.69 $\pm$ \small{0.03} & 5.25 $\pm$ \small{0.46} & 9.15 $\pm$ \small{0.18} & 8.04 $\pm$ \small{0.29} & 8.48 $\pm$ \small{0.43} & 8.48 $\pm$ \small{0.43} & 6.64 $\pm$ \small{0.06} & 6.45 $\pm$ \small{0.06} \\
			MAS + FL & 7.90 $\pm$ \small{0.33} & 5.34 $\pm$ \small{0.11} & 6.01 $\pm$ \small{0.16} & 2.12 $\pm$ \small{0.27} & 2.34 $\pm$ \small{0.14} & 2.79 $\pm$ \small{0.30} & 2.12 $\pm$ \small{0.01} & 2.12 $\pm$ \small{0.01} & 4.59 $\pm$ \small{0.16} & 3.09 $\pm$ \small{0.17} \\
			LwF + FL & 10.95 $\pm$ \small{0.65} & 2.46 $\pm$ \small{0.09} & \textbf{17.00} $\pm$ \small{0.35} & 6.03 $\pm$ \small{0.21} & 15.10 $\pm$ \small{1.09} & 13.75 $\pm$ \small{2.11} & 15.88 $\pm$ \small{1.23} & 15.88 $\pm$ \small{1.23} & 14.73 $\pm$ \small{0.40} & 9.53 $\pm$ \small{0.19} \\
			FCIL & 10.69 $\pm$ \small{0.78} & \textbf{7.51} $\pm$ \small{0.03} & 14.90 $\pm$ \small{1.60} & 10.37 $\pm$ \small{0.79} & 14.49 $\pm$ \small{1.74} & \textbf{13.84} $\pm$ \small{0.10} & 14.34 $\pm$ \small{0.78} & 14.34 $\pm$ \small{0.78} & 13.60 $\pm$ \small{0.44} & 11.52 $\pm$ \small{0.36} \\ \hline 
			\textbf{\ourapproach} & 10.48 $\pm$ \small{0.58} & 7.31 $\pm$ \small{0.32} & 13.52 $\pm$ \small{0.43} & \textbf{10.41} $\pm$ \small{0.79} & \textbf{16.17} $\pm$ \small{1.47} & 13.75 $\pm$ \small{0.44} & \textbf{19.93} $\pm$ \small{0.26} & \textbf{19.93} $\pm$ \small{0.26} & \textbf{15.03} $\pm$ \small{0.40} & \textbf{12.85} $\pm$ \small{0.24} \\ \hline
		\end{tabular}
	}
	\caption{The test accuracy (\%) of each task at the round of task switch (@TS), and after the final round(@Fin), of the class-incremental settings.}
	\label{tab:ci_res_dev}
\end{sidewaystable*}

\begin{sidewaystable*}
	\centering
	\resizebox{1.\linewidth}{!}{
		\begin{tabular}{|c|c|c|c|c|c||c|c|c|c|}
			\hline
			\multicolumn{2}{|c|}{} & \multicolumn{4}{c||}{\textbf{Small-Scale Image Classification}} &  \multicolumn{4}{c|}{\textbf{Large-Scale Image Classification}}\\ \hline
			\multicolumn{2}{|c|}{Methods} &  Task 1 & Task 2 & Task 3 &  Avg. &  Task 1 & Task 2 & Task 3 &  Avg. \\ \hline
			\multirow{2}{*}{FedAvg} & @TS & 10.94 $\pm$ \small{0.84} & 13.84 $\pm$ \small{0.71} & 5.25 $\pm$ \small{0.44} & 10.01 $\pm$ \small{0.20} & 29.13 $\pm$ \small{0.68} & 10.88 $\pm$ \small{0.16} & 5.13 $\pm$ \small{0.12} & 15.05 $\pm$ \small{0.32} \\ \cline{2-10}
			& @Fin & 10.60 $\pm$ \small{2.45} & 4.24 $\pm$ \small{0.41} & 5.25 $\pm$ \small{0.44} & 6.70 $\pm$ \small{0.54} & 9.25 $\pm$ \small{0.12} & 3.98 $\pm$ \small{0.13} & 5.13 $\pm$ \small{0.12} & 6.12 $\pm$ \small{0.05} \\ \hline
			\multirow{2}{*}{SCAFFOLD} & @TS & 26.49 $\pm$ \small{0.56} & 4.58 $\pm$ \small{0.16} & 4.00 $\pm$ \small{0.08} & 11.69 $\pm$ \small{0.26} & \textbf{37.50} $\pm$ \small{1.45} & 8.25 $\pm$ \small{0.17} & 6.00 $\pm$ \small{0.07} & 17.25 $\pm$ \small{0.56} \\ \cline{2-10}
			& @Fin & 21.92 $\pm$ \small{0.54} & 3.68 $\pm$ \small{0.17} & 4.00 $\pm$ \small{0.08} & 9.87 $\pm$ \small{0.25} & 27.38 $\pm$ \small{2.19} & 7.75 $\pm$ \small{0.05} & 6.00 $\pm$ \small{0.07} & 13.71 $\pm$ \small{0.76} \\ \hline
			\multirow{2}{*}{EWC+FL} & @TS & 25.40 $\pm$ \small{0.17} & 2.23 $\pm$ \small{0.19} & 3.78 $\pm$ \small{0.17} & 10.47 $\pm$ \small{0.16} & 32.00 $\pm$ \small{1.43} & 12.75 $\pm$ \small{1.21} & 4.75 $\pm$ \small{0.13} & 16.50 $\pm$ \small{0.91} \\ \cline{2-10}
			& @Fin & \textbf{22.54} $\pm$ \small{1.78} & 2.23 $\pm$ \small{0.21} & 3.78 $\pm$ \small{0.17} & 9.52 $\pm$ \small{0.70} & 16.50 $\pm$ \small{0.99} & 6.12 $\pm$ \small{0.21} & 4.75 $\pm$ \small{0.13} & 9.12 $\pm$ \small{0.34} \\ \hline
			\multirow{2}{*}{MAS+FL} & @TS & 11.16 $\pm$ \small{0.65} & 11.45 $\pm$ \small{0.28} & 4.56 $\pm$ \small{0.11} & 9.06 $\pm$ \small{0.33} & 12.75 $\pm$ \small{0.84} & 10.88 $\pm$ \small{0.99} & 3.50 $\pm$ \small{0.17} & 9.04 $\pm$ \small{0.58} \\ \cline{2-10}
			& @Fin & 10.60 $\pm$ \small{0.87} & 5.89 $\pm$ \small{0.15} & 4.56 $\pm$ \small{0.11} & 7.02 $\pm$ \small{0.31} & 11.50 $\pm$ \small{0.65} & 4.00 $\pm$ \small{0.22} & 3.50 $\pm$ \small{0.17} & 6.33 $\pm$ \small{0.15} \\ \hline
			\multirow{2}{*}{LwF+FL} & @TS & 34.17 $\pm$ \small{1.38} & \textbf{15.78} $\pm$ \small{0.59} & 8.26 $\pm$ \small{0.18} & 19.40 $\pm$ \small{0.59} & 33.50 $\pm$ \small{2.08} & \textbf{17.87} $\pm$ \small{0.14} & \textbf{6.13} $\pm$ \small{0.24} & 19.17 $\pm$ \small{0.68} \\ \cline{2-10}
			& @Fin & 11.27 $\pm$ \small{1.06} & 5.25 $\pm$ \small{0.40} & 8.26 $\pm$ \small{0.18} & 8.26 $\pm$ \small{0.48} & 11.87 $\pm$ \small{0.93} & 4.87 $\pm$ \small{0.16} & \textbf{6.13} $\pm$ \small{0.24} & 7.62 $\pm$ \small{0.30} \\ \hline \hline
			\multirow{2}{*}{\textbf{\ourapproach}} & @TS & \textbf{44.45} $\pm$ \small{1.02} & 13.37 $\pm$ \small{0.73} & \textbf{8.45} $\pm$ \small{0.15} & \textbf{22.09} $\pm$ \small{0.53} & 37.28 $\pm$ \small{1.55} & 14.58 $\pm$ \small{0.91} & 5.91 $\pm$ \small{0.32} & \textbf{19.26} $\pm$ \small{0.11} \\ \cline{2-10}
			& @Fin & 22.23 $\pm$ \small{1.24} & \textbf{13.09} $\pm$ \small{1.03} & \textbf{8.45} $\pm$ \small{0.15} & \textbf{14.59} $\pm$ \small{0.71} & \textbf{31.83} $\pm$ \small{1.46} & \textbf{12.80} $\pm$ \small{2.34} & 5.91 $\pm$ \small{0.32} & \textbf{16.85} $\pm$ \small{1.30} \\ \hline
		\end{tabular}
	}
	\caption{The test accuracy (\%) of each task at the round of task switch (@TS), and after the final round(@Fin), of the task-incremental settings.}
	\label{tab:ti_res_dev}
\end{sidewaystable*}

\begin{sidewaystable*}
	\centering
	\resizebox{1.\linewidth}{!}{
		\begin{tabular}{|c|c|c|c|c|c||c|c|c|c|}
			\hline
			& \multicolumn{5}{c||}{\textbf{Class-Incr, Tiny-ImageNet}} &  \multicolumn{4}{c|}{\textbf{Task-Incr, Image Classification}}\\ \hline
			Methods &  Task 1 & Task 2 & Task 3 & Task 4 &  Avg. &  Task 1 & Task 2 & Task 3 &  Avg. \\ \hline
			FedAvg & 2.12 $\pm$ \small{0.15} & 2.01 $\pm$ \small{0.14} & 2.34 $\pm$ \small{0.07} & 2.57 $\pm$ \small{0.09} & 2.26 $\pm$ \small{0.07} & 12.05 $\pm$ \small{0.99} & 3.01 $\pm$ \small{0.25} & 6.70 $\pm$ \small{0.12} & 7.25 $\pm$ \small{0.29} \\ \hline
			SCAFFOLD & 8.04 $\pm$ \small{0.32} & 11.94 $\pm$ \small{1.96} & 12.39 $\pm$ \small{0.76} & 19.42 $\pm$ \small{1.22} & 12.95 $\pm$ \small{0.89} & 37.21 $\pm$ \small{0.82} & 5.02 $\pm$ \small{0.25} & 1.56 $\pm$ \small{0.34} & 14.60 $\pm$ \small{0.45} \\ \hline
			EWC+FL & 5.47 $\pm$ \small{0.08} & 7.59 $\pm$ \small{0.38} & 7.92 $\pm$ \small{0.13} & 12.72 $\pm$ \small{0.80} & 8.43 $\pm$ \small{0.23} & 17.86 $\pm$ \small{0.34} & 2.34 $\pm$ \small{0.08} & 1.12 $\pm$ \small{0.26} & 7.11 $\pm$ \small{0.17} \\ \hline
			MAS+FL & 2.34 $\pm$ \small{0.06} & 2.23 $\pm$ \small{0.09} & 2.34 $\pm$ \small{0.18} & 2.57 $\pm$ \small{0.54} & 2.37 $\pm$ \small{0.16} & 11.05 $\pm$ \small{1.43} & 1.12 $\pm$ \small{0.28} & 0.89 $\pm$ \small{0.13} & 4.35 $\pm$ \small{0.48} \\ \hline
			LwF+FL & 6.03 $\pm$ \small{0.06} & 11.83 $\pm$ \small{0.35} & \textbf{24.33} $\pm$ \small{1.49} & 27.23 $\pm$ \small{0.02} & 17.36 $\pm$ \small{0.35} & 15.96 $\pm$ \small{0.60} & 8.48 $\pm$ \small{0.33} & \textbf{8.93} $\pm$ \small{0.22} & 11.12 $\pm$ \small{0.25} \\ \hline
			\textbf{\ourapproach} & \textbf{8.04} $\pm$ \small{0.33} & \textbf{18.28} $\pm$ \small{1.34} & 21.45 $\pm$ \small{2.20} & \textbf{30.34} $\pm$ \small{1.39} & \textbf{19.53} $\pm$ \small{0.46} & \textbf{42.56} $\pm$ \small{1.20} & \textbf{15.23} $\pm$ \small{2.15} & 8.29 $\pm$ \small{0.10} & \textbf{22.03} $\pm$ \small{0.80} \\ \hline
		\end{tabular}
	}
	\caption{The final accuracy (\%) under gradual FCL settings.}
	\label{tab:grad_res_dev}
\end{sidewaystable*}

\subsection{Effect of Participation Ratio}
In the main paper, we use 100 clients in total, and selects 10 clients at each round for training, i.e. the participation ratio is set to 0.1. To better understand the effect of different participation ratios, we scale the participation ratio in between 0.1 and 0.6 and conduct further experiments. Specifically, we choose 10, 20, 40, 60 clients out of total 100 clients to obtain results with participation ratio 0.1, 0.2, 0.4, 0.6. We run the experiment under the class-incremental Tiny-ImageNet setting and the task-incremental small-scale image-classification setting.

\begin{figure*}[ht]
	\centering
	\begin{subfigure}[t]{0.48\linewidth}
		\centering
		\includegraphics[width=\textwidth]{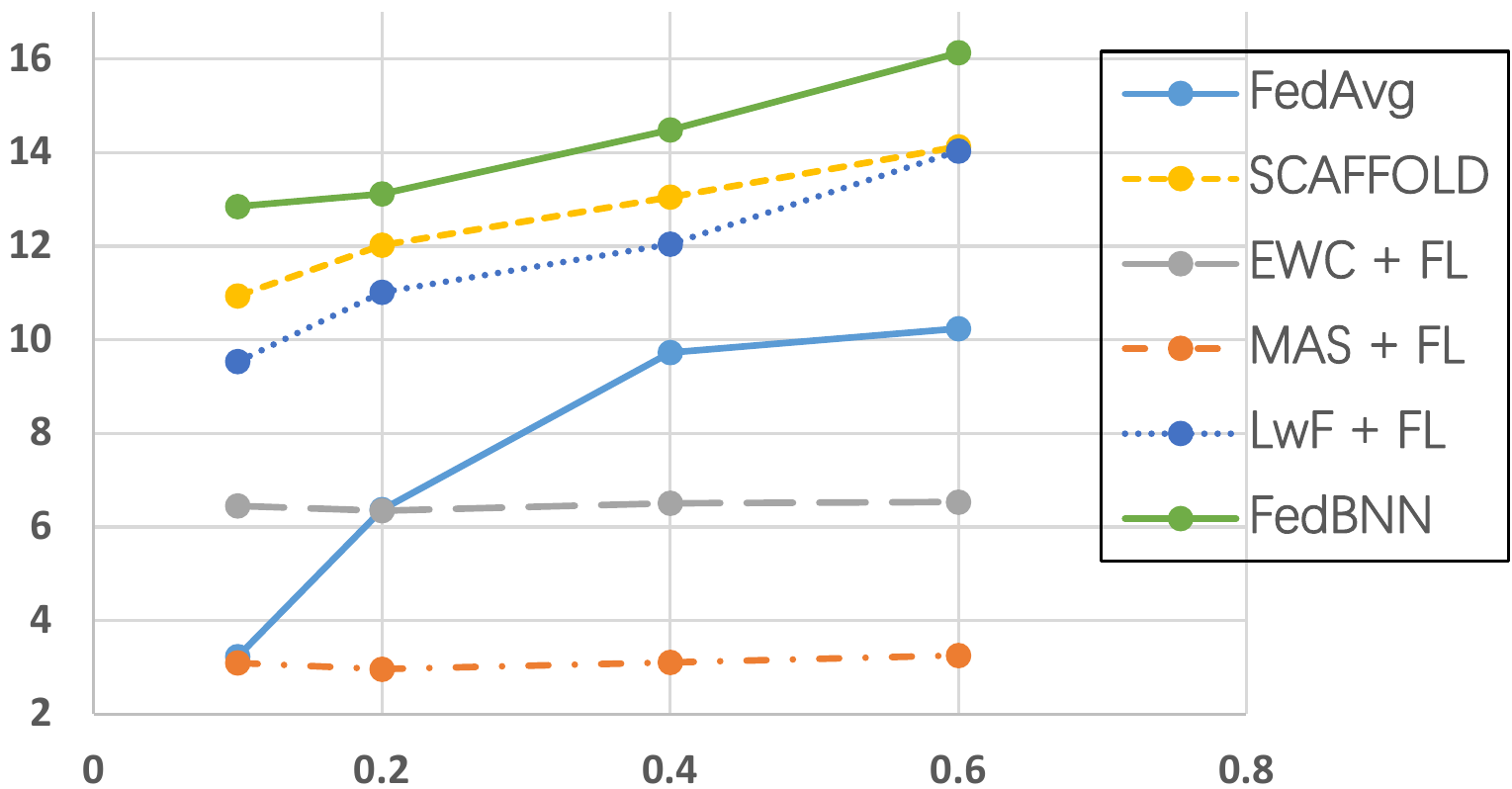}
		\caption{Average final accuracy of class-incremental FCL setting on Tiny-ImageNet. X-axis is the participation ratio, and Y-axis is accuracy (\%). }
		\label{fig:timnet_ci_part}
	\end{subfigure}
	\hspace{0.02\linewidth}
	\begin{subfigure}[t]{0.48\linewidth}
		\centering
		\includegraphics[width=\textwidth]{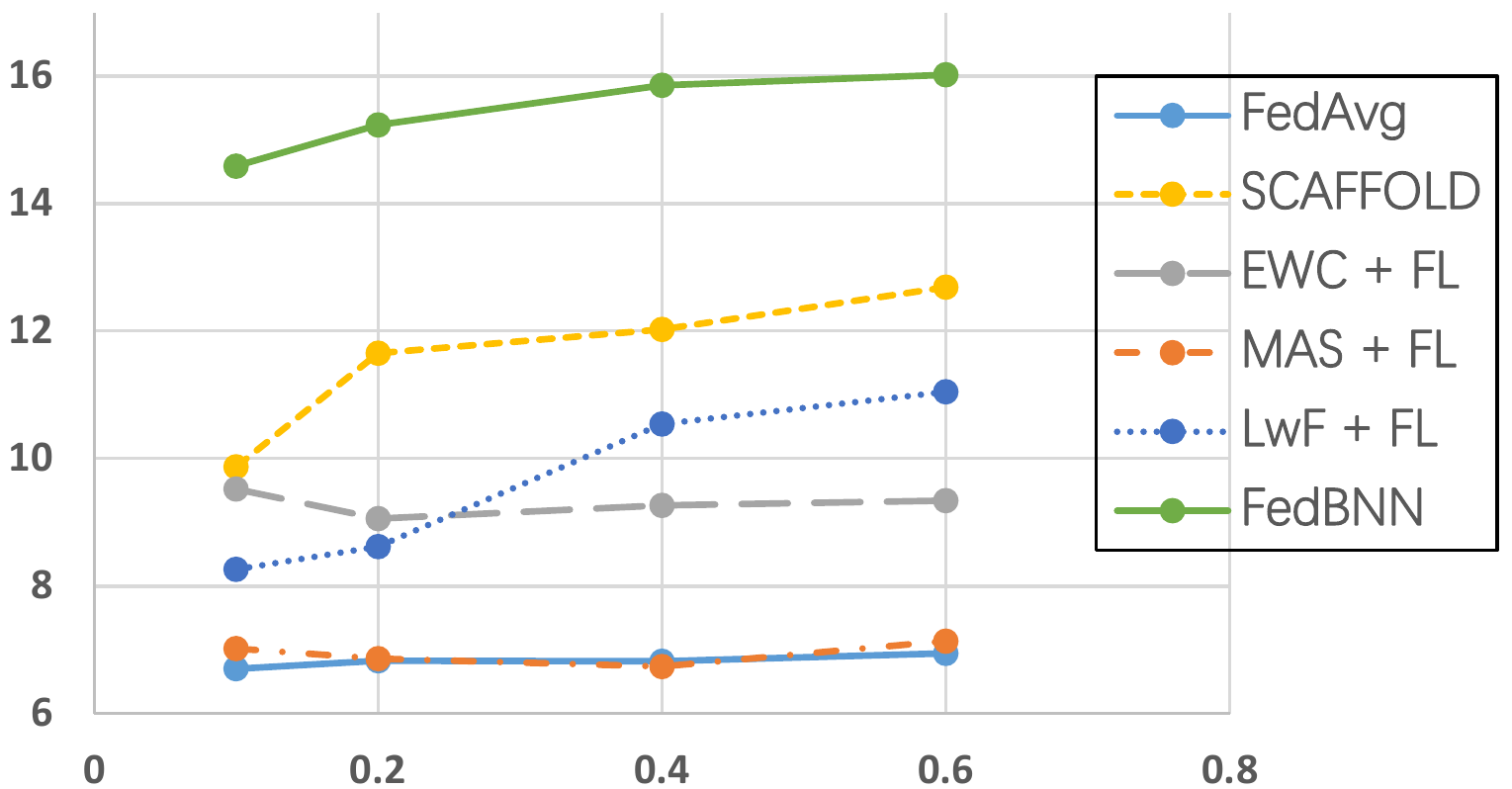}
		\caption{Average final accuracy of task-incremental FCL setting on small-scale image classification. X-axis is the participation ratio, and Y-axis is accuracy (\%).}
		\label{fig:oc1_ti_part}
	\end{subfigure}
	\caption{Average final accuracy of the algorithms with varying participation ratio. }
	\label{fig:partial_participation}
\end{figure*}

\cref{fig:partial_participation} presents the accuracy of the compared baselines with regard to different participation ratios. It can be observed that while some baselines (MAS + FL and EWC + FL) are less sensitive to participation ratio, most algorithms see improvements in terms of accuracy with increasing participation ratio. Furthermore, \ourapproach consistently outperform the baselines with different participation ratios, which demonstrates the effectiveness of our approach.

\end{document}